\documentclass[journal]{IEEEtran}
\ifCLASSINFOpdf
\else
\fi
\usepackage{tabu}
\usepackage{color}
\usepackage{array}
\usepackage{footnote}
\usepackage{mathptmx}
\usepackage{cite}
\usepackage{upgreek}
\usepackage{color}
\usepackage{gensymb}
\usepackage{mathrsfs}
\usepackage{siunitx}
\usepackage{algorithm}
\usepackage{algpseudocode}
\usepackage{graphicx}

\usepackage{textcomp}
\usepackage{amsmath}
\usepackage{amsfonts}
\usepackage{graphicx}
\usepackage{float}
\usepackage{subfig}
\graphicspath{{figs/}}
\usepackage[labelfont=bf]{caption}
\usepackage[utf8]{inputenc}
\usepackage{mathtools}
\usepackage{booktabs}

\DeclareMathOperator*{\argminB}{argmin}

\usepackage{amsmath}
\usepackage{url}

\begin{document}
%
\title{Geometric Solutions for General Actuator Routing on Inflated-Beam Soft Growing Robots}
%
%
%

\author{Laura~H.~Blumenschein,~\IEEEmembership{Member,~IEEE,}
        Margaret~Koehler,~\IEEEmembership{Student~Member,~IEEE,}
        Nathan~S.~Usevitch,~\IEEEmembership{Student~Member,~IEEE,}
        Elliot~W.~Hawkes,~\IEEEmembership{Member,~IEEE,} \\
        D.~Caleb~Rucker,~\IEEEmembership{Member,~IEEE,}
        and~Allison~M.~Okamura,~\IEEEmembership{Fellow,~IEEE}
\thanks{\copyright 2021 IEEE. Personal use of this material is permitted. Permission from IEEE must be obtained for all other uses, in any current or future media, including reprinting/republishing this material for advertising or promotional purposes, creating new collective works, for resale or redistribution to servers or lists, or reuse of any copyrighted component of this work in other works.\newline \indent Manuscript received May 21, 2021; accepted August 22, 2021. This work was supported by National Science Foundation Awards \#1637446, \#1652588, and \#2024247, Air Force Office of Scientific Research Award FA2386-17-1-4658, the National Science Foundation Graduate Fellowship Program, and Toyota Research Institute (TRI). TRI provided funds to assist the authors with their research, but this article solely reflects the opinions and conclusions of its authors and not TRI or any other Toyota entity. This article was recommended for publication by Associate
Editor  and Editor M. Yim upon evaluation of the reviewers’
comments. \textit{(Corresponding author: Laura Blumenschein)}}
\thanks{L. H. Blumenschein is with the School of Mechanical Engineering, Purdue University, West Lafayette, IN 47907 USA; \newline {\tt\small lhblumen@purdue.edu}.\newline \indent
M. Koehler, N. S. Usevitch, and A. M. Okamura are with the Department of Mechanical Engineering, Stanford University, Stanford, CA 94035 USA \newline {\tt\small aokamura@stanford.edu}.\newline \indent 
    E. W. Hawkes is with the Department of
Mechanical Engineering, University of California, Santa Barbara, CA 93106 USA \newline {\tt\small ewhawkes@engineering.ucsb.edu}.\newline \indent 
    D. C. Rucker is with the Department of Mechanical, Aerospace and Biomedical Engineering, University of Tennessee, Knoxville TN 37996 USA  \newline {\tt\small drucker6@ibme.utk.edu}. \newline \indent
 Digital Object Identifier 10.1109/TRO.2021.3115230}
}

%
%

\markboth{IEEE Transactions on Robotics}%
{Blumenschein \MakeLowercase{\textit{et al.}}: Geometric Solutions for General Actuator Routing on Inflated-Beam Soft Growing Robots}
%



\maketitle

\begin{abstract}
Continuum and soft robots can leverage complex actuator shapes to take on useful shapes while actuating only a few of their many degrees of freedom. Continuum robots that also grow increase the range of potential shapes that can be actuated and enable easier access to constrained environments. Existing models for describing the complex kinematics involved in general actuation of continuum robots rely on simulation or well-behaved stress-strain relationships, but the non-linear behavior of the thin-walled inflated-beams used in growing robots makes these techniques difficult to apply. Here we derive kinematic models of single, generally routed tendon paths on a soft pneumatic backbone of inextensible but flexible material from geometric relationships alone. This allows for forward modeling of the resulting shapes with only knowledge of the geometry of the system. We show that this model can accurately predict the shape of the whole robot body and how the model changes with actuation type. We also demonstrate the use of this kinematic model for inverse design, where actuator designs are found based on desired final robot shapes. We deploy these designed actuators on soft pneumatic growing robots to show the benefits of simultaneous growth and shape change.
\end{abstract}


%
\IEEEpeerreviewmaketitle

\section{Introduction}
Compliance in soft or continuum robots allows them to take on a wide variety of shapes \cite{Rus2015a,webster2010design}. Continuum robots are often said to have ``infinite" passive degrees of freedom, any of which can potentially be actuated. Actuators that leverage these continuous degrees of freedom in interesting ways enable numerous compelling behaviors and applications. In general, well-informed design of actuation strategies for soft robots requires knowledge of the relationship between the actuator design and the resulting kinematics. Helical tendon routing, for example, can be used to expand the reachable workspace of a manipulator \cite{starke2017merits} or create multibend shapes \cite{gao2015cross}. While some types of continuum robots have well-described kinematics and dynamics, existing solutions cannot always be easily extended to soft robots with compressible backbones or non-traditional backbone geometries and materials, properties shared by many recent growing continuum robots~\cite{hawkes2017,coad2019vine,putzu2018plant,abrar2019epam}. In this work, we present a geometric model for one such compressible backbone continuum robot and use the model to develop the kinematics of a generally routed actuator on a thin-walled, soft pneumatic growing robot.

In general, the flexible and elastic materials that make up soft robots can be continuously deformed into a wide variety of shapes. Soft robots have been developed that bend \cite{polygerinos2013towards,mazzolai2012soft,mcmahan2005design}, twist \cite{bishop2015design,uppalapati2018towards}, extend \cite{Hawkes2016}, expand \cite{steltz2009jsel}, and carry out complex motions \cite{connolly2015mechanical, gilbertson2017serially} with only a few actuator inputs. Complex motions of soft robots have been used to grasp objects \cite{mazzolai2012soft,mcmahan2005design}, locomote \cite{shepherd2011multigait}, make haptic displays \cite{stanley2016closed}, and more \cite{Rus2015a}. The shapes and motions of soft robots are dependent on both the make-up of the soft body and the coupling of the body and actuation, so models of these relationships are useful for designing the actuation. 

\begin{figure}[t!]
\begin{center}
	\includegraphics[width=.95\columnwidth]{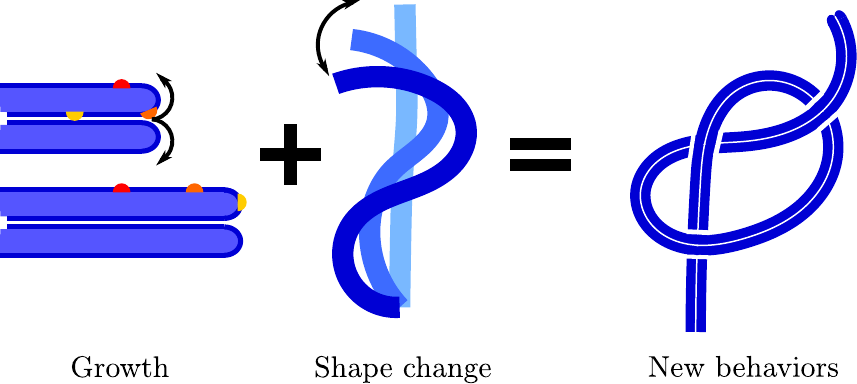}
	\caption{\small Inflated-beam soft growing robots can move through two different means, growing from the tip and changing their shape. These degrees of freedom work independently of each other and, when combined, yield new behaviors.}
	\vspace{-0.2cm}
	\label{fig:overview}
    \end{center}
\end{figure}

In this paper we focus on one particular soft robot that uses a flexible but inextensible pressurized tube to form a pneumatic backbone, and has a long flexible body similar in form to many continuum robots \cite{hawkes2017,greer2019soft}. This pneumatic backbone robot effectively ``grows" through everting material, similar to whole skin locomotion \cite{ingram2005whole}, in order to move into and explore its environment \cite{coad2019vine} and can create tight curvatures and spirals \cite{blumenschein2018helical,gan20203d}. Growth as a form of movement allows for easy traversal through constrained environments and the formation of a useful structure along the grown path \cite{hawkes2017, coad2019vine}, but it requires a thin-walled pneumatic backbone in order to allow material eversion at the tip. This thin-walled pneumatic backbone results in non-linear local buckling and wrinkling behaviors that are not easily modeled using traditional continuum robot techniques. Growing robots have unique benefits as continuum robots (Fig.~\ref{fig:overview}); in particular, growth and shape actuation (steering) represent orthogonal and independent degrees of freedom, so general shape change can be investigated without considering growth. Adding growth to shape change also expands the potential reachable space of shapes using only a single steering actuator. Modeling and control for growing continuum robots has provided methods for tip position and orientation \cite{greer2019soft,el2019dynamic,ataka2020model}, but models of the full robot shape have not previously been considered.

Kinematics and mechanics models for soft robot deformation have used many techniques, from geometric relationships \cite{jones2006kinematics} to finite element methods \cite{coevoet2019soft}. For continuum robots, we are specifically interested in applying these techniques for lines of actuation placed in a general shape around the continuum robot backbone, henceforth referred to as general actuator routing or general tendon routing. Constant curvature models are able to use geometric constraints of the robot to predict actuation of tendons parallel to the backbone \cite{jones2006kinematics,webster2010design}, but these specific geometric relationships used apply narrowly to only those parallel actuators. By adding mechanics to geometric constraints, more general models for continuum robots have been created. Continuum robot models that predict deformations under general tendon routing and external loading have been built on Cosserat rod and Cosserat string methods \cite{rucker2011,Till2019,Butler2019}. These models generally require knowledge of both the geometry and the stress-strain relationship of the backbone. For thin-walled pneumatic backbones, the stress-strain relationship is highly nonlinear and difficult to model due to the buckling behavior of the thin-walled tube \cite{fichter1966theory,comer1963deflections}, which makes it difficult to adapt these Cosserat-based models. Looking beyond continuum robot models, the models for fiber-reinforced elastomeric enclosures (FREEs) \cite{uppalapati2018towards,krishnan2015kinematics} and McKibben actuators \cite{tondu2000modeling} are geometrically similar to the models of continuum robots and rely on distributed inextensibility constraints from the wound fibers. The deformation of FREEs has been modeled both geometrically, based on the inextensible fibers, and through finite element methods. The finite element methods are often more accurate as they can account for non-idealities that the geometric models ignore, but they have difficulty accounting for highly non-linear material behaviors like buckling or wrinkling \cite{krishnan2015kinematics}. While these methods can inspire models for thin-walled pneumatic continuum robots, the particular geometric relationships exploited do not transfer. In this work, we define the geometric constraints to develop a kinematic model for general tendon routing on thin-walled pneumatic continuum robots and show how these constraints can be used to model the deformation of the pneumatic tube under various types of tendon-like actuation.

The remainder of the paper is organized as follows. Section~\ref{sec:ch5_geomModel} presents the geometric solution for helical actuation and shows how to extend this solution to general paths. We then present different methods for actuating a soft growing robot and the benefits and limitations of each actuation method in terms of the model (Section~\ref{sec:ch5_actImplementation}). Section~\ref{sec:ch5_modelValidation} presents experimental results validating first the uniform helical actuation and then the general form of the model. Section~\ref{sec:ch5_invDesign} shows another use for this model, to design actuators to achieve desired shapes. Lastly, we end with a discussion of the benefits of growth for general actuation and the limitations of the actuation methods presented. Parts of this work were previously published in \cite{blumenschein2018helical}, which initially presented the model for uniform helical actuation, descriptions of actuator implementations, and initial results validating the helical model. This paper builds on that initial work, with the following new contributions: generalizing the model from helical tendons to any tendon shape, validating both the helical and general actuator models with new experiments, and describing a method to design actuators to match target shapes.

\section{Geometric Model}
\label{sec:ch5_geomModel}
To develop a geometric model for thin-walled pneumatic growing robots, we first define the problem for uniform routings of actuators, i.e.\ helical actuation, including the actuator parameterization and geometric constraints, and present the closed-form solution for helical actuator kinematics. We then show how this closed-form solution for a uniform helical path can be used to calculate the actuation of general paths. While the function of growth defines the structure of the robot, previous work has shown that steering and growth can be treated as independent degrees of freedom \cite{greer2019soft}, so we do not need to explicitly consider growth in the geometric model.

\subsection{Uniform Actuation Geometry}
\subsubsection{Actuator Geometry}

\begin{figure}[t!]
\begin{center}
	\includegraphics[width=\columnwidth]{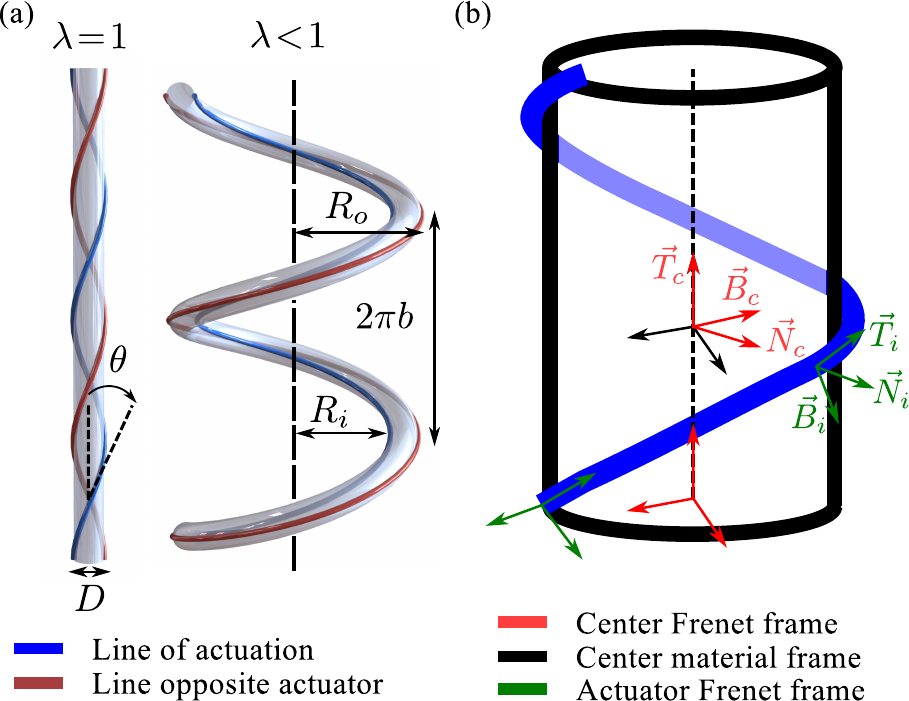}
	\caption[Visualization of different types of uniform cable routings and their effect on robot shape, with the actuated cable shown in blue.]{\small Visualization of uniform cable routings and their effect on robot shape, with the actuated cable shown in blue and the diametrically opposite line in red. In (a), the robot with helical cables is shown in both unactuated and actuated states. The actuator parameters, $\theta$, $D$, $\lambda$, and shape parameters, $R_o$, $R_i$, $b$, are displayed. In (b), the conventions for the Frenet-Serret frames along the actuator path and the Frenet-Serret and material frames along center-line path in the unactuated case are shown.}
	\label{fig:ch5_Params}
    \end{center}
\end{figure}

To achieve a uniform deformed shape, the continuum robot actuator must route in a uniform path on the surface of the body. Here we use uniform to mean that any segment of the shape is geometrically similar to all other segments of the shape. Uniform tendons may be routed on the robot's surface axially or at an angle, forming a helix on the body of the tube (Fig.~\ref{fig:ch5_Params}(a)). The traditional routing of tendons on a continuum robot, in a straight path parallel to the undeformed backbone, leads to a geometrically self-similar shape in 2D when actuated, i.e. constant-curvature deformation \cite{webster2010design}. Helical tendon routing similarly leads to a symmetrical and self-similar shape in 3D, resulting in a helical actuated shape (Fig.~\ref{fig:ch5_Params}(a)). This behavior has been previously observed for a soft pneumatic continuum robot in practice in our previous work \cite{blumenschein2018helical} and closely resembles the methods natural growing systems, like plants, use to form helical structures \cite{smyth2016helical}.

We will quantify the relationship between the helical path of an actuator around an undeformed pneumatic tube and the resulting helical shape of the actuated pneumatic tube. To develop this relationship, we first give the standard parameterization of the 3D path of a helix, $\vec{r}(s)$, in terms of its radius and pitch,
\begin{equation}
\vec{r}(s)= \begin{bmatrix} R\sin(s)  & R\cos(s) & b \: s \end{bmatrix}^T
\label{eqn:Helix}
\end{equation}
where $R$ is the radius of a helical path, $b$ is the normalized pitch parameter such that $2\pi b$ is the height achieved by one revolution of the helix, and $s\in [0,S]$ is the length based variable for parameterization, where the total length is $S\sqrt{b^2+R^2}$. This helix parameterization applies both to the robot shape and the actuator shape. 

For the actuator, the helical path must lie on the body of the robot, so the actuator radius is equal to the thin-walled tube radius. In our parameterization of the helical actuator path we use variations of these standard parameters. For the intuitiveness of our formulation and to allow our model to cover the straight actuator case as well, it is convenient to use the tube diameter, $D$, in place of the radius and to replace the actuator pitch, $b_a$, with the drawn angle of the path with respect to a straight actuator, $\theta$, which is defined as
\begin{equation}
\theta = \arctan \frac{D}{2 b_{a}}.
\label{eqn:theta}
\end{equation}
\par

These parameters define the path of the uncontracted actuator. A third parameter is necessary to describe the amount of actuation along the helical path. For a purely geometric model, we parameterize the actuation using the relative shortening of the path length when actuated compared to the path length when not actuated. We use $\lambda$ to represent this ratio. The buckling behavior of the flexible but inextensible pneumatic tube wall means that how we use this actuator parameter is highly dependent on how the tube is actuated. The contraction ratio, $\lambda$, can be physically achieved in several different ways, as detailed in Section~\ref{sec:ch5_actImplementation}, which can generally be categorized as either continuous, meaning a single actuator can achieve values for $\lambda$ within a range, or discrete, meaning an actuator switches between a few set values for $\lambda$. These considerations and specifications of the implemented actuation will be discussed in detail in Section~\ref{sec:ch5_actImplementation}. Fig.~\ref{fig:ch5_Params}(a) shows the parameters used to describe the actuator shape on the initial tube and the parameters of the resulting helix after actuation.

\subsubsection{Deformed Robot Geometry}

To understand the resulting shape further, we observe how the inflated-beam soft robot deforms to achieve the final shape. The pneumatic beam is made of inextensible plastic or fabric so it can only shorten to change shape, not lengthen. It accomplishes this length change by wrinkling the thin wall of the material. In some implementations of actuation, this occurs at discrete locations defined by the construction of the actuator (Section~\ref{sec:ch5_actImplementation}). The distributed strain caused either by the distributed loading or the mechanical constraints, depending on the actuator implementation, causes distributed wrinkling that approximates a continuous shortening along the path of the actuator. We assume that the maximum wrinkling will occur at the location of the actuator, and no wrinkling will occur diametrically opposite to the point of maximum wrinkling. 

We define the robot shape by parameterizing the actuator path $\vec{r}_i(s)$ (the path with the highest wrinkling) and the path diametrically opposite the actuator $\vec{r}_o(s)$ (the path with no wrinkling) with a common parameter $s$ (Fig. \ref{fig:GeoCon}) as follows: 

\begin{equation}
\vec{r}_i(s)=\begin{bmatrix} R_i\sin(s)  & R_i\cos(s) & b \: s \end{bmatrix}^T 
\label{eq:r_i}
\end{equation}
\begin{equation}
\vec{r}_o(s)=\begin{bmatrix} R_o\sin(s)  & R_o\cos(s) & b \: s \end{bmatrix}^T 
\label{eq:r_o}
\end{equation}
where $R_i$ is the radius of the inner helix, $R_o$ is the radius of the outer helix, and $b$ is the normalized pitch of the helices (Fig.~\ref{fig:ch5_Params}(a)). The inner and outer paths share the same pitch since they are attached to a single tube and can not diverge from each other.

In addition to the paths on the surface of the tube, we also define the center path of the robot as:

\begin{equation}
\vec{r}_c(s)=\frac{1}{2}(\vec{r}_i(s)+\vec{r}_o(s)).
\label{eqn:ch5_centerPath}
\end{equation}

\subsection{Geometric Constraints}
\begin{figure}[t!]
\begin{center}
	\includegraphics[width=.9\columnwidth]{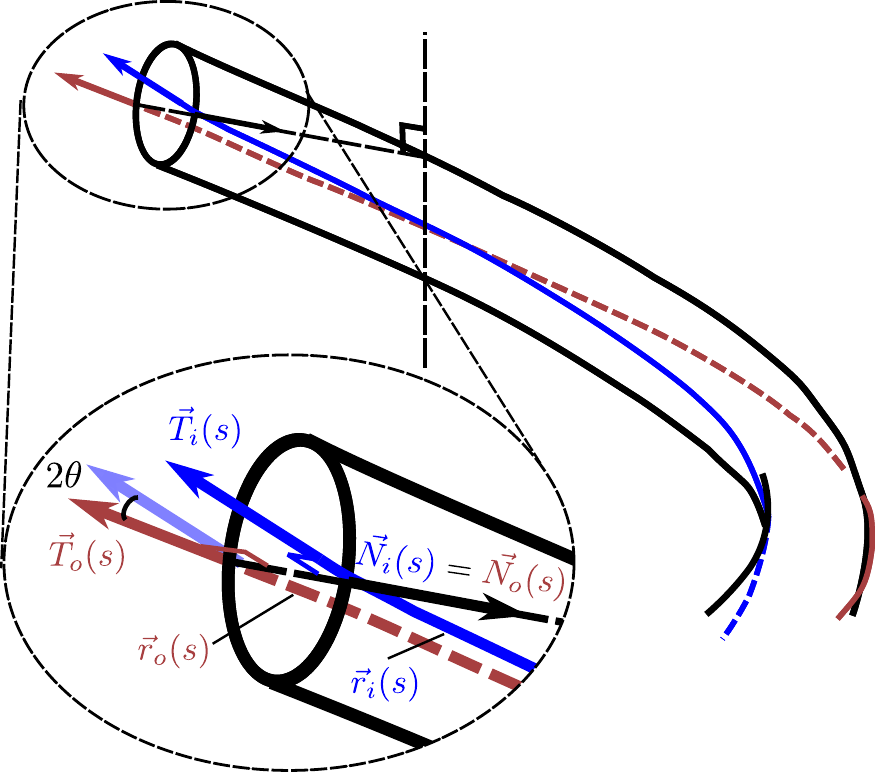}
	\caption[Diagram of the geometric relationships between the actuator path (blue) and the diametrically opposite path (red).]{\small Diagram of the geometric relationships between the actuator path (blue) and the diametrically opposite path (red). The two paths are parameterized by $s$. The points $\vec{r}_o(s)$ and $\vec{r}_i(s)$ are separated by a distance $D$. The vector $\vec{r}_o(s)-\vec{r}_i(s)$ is coincident with the normal vectors ($\vec{N}_o(s)$ and $\vec{N}_i(s)$) and orthogonal to the tangent vectors to these curves ($\vec{T}_o(s)$ and $\vec{T}_i(s)$). The angle between the tangent vectors is $2\theta$, or twice the drawn angle of the actuator relative to the straight tube.}
    \label{fig:GeoCon}
\end{center}
\end{figure}

The relationship between the parameters defining the unactuated path and actuation state of the actuator ($\theta$, $D$, $\lambda$) and the parameters defining the final shape of an actuated helix ($R_o$, $R_i$, $b$) are developed through an understanding of the geometric constraints imposed by the tube material. We define constraints to relate the shortest and longest helical paths, $\vec{r}_i(s)$ and $\vec{r}_o(s)$ respectively, to each other using the actuator parameters.

Looking at a given length of the helix-actuated inflated tube, we know that the ratio of the inner arc length to the outer arc length over any portion of the robot should be equal to the contraction ratio, $\lambda$. For the helices described in Equations~(\ref{eq:r_i}) and (\ref{eq:r_o}), and looking at the interval  $s \in [0,S]$, this constraint can be written:
\begin{equation}
\lambda =\frac{\sqrt[]{(Sb)^2+ (SR_i)^2}}{\sqrt[]{(Sb)^2+ (SR_o)^2}}=\frac{\sqrt[]{b^2+ R_i^2}}{\sqrt[]{b^2+ R_o^2}}.
\label{eqn:act_len_constraint}
\end{equation}
This constraint is defined by how much the inner path is shortened relative to the outer, unwrinkled, path. 

The remaining constraints define the relationship between the inner and outer paths for any points $\vec{r}_i(s)$ and $\vec{r}_o(s)$ for a common value $s$ along the robot. These constraints are portrayed graphically in Fig.~\ref{fig:GeoCon}. When the system is under pressure, the material is in tension and the volume is maximized. In the unactuated state, the robot is a tube and the cross-sections normal to the centerline path, $\vec{r}_c(s)$ for $(s \in [0, S])$, are circles of diameter $D$. As the tube is actuated and wrinkles, we observe that these cross-sections move relative to one another but remain approximately circular, and due to the relative inextensibility of the materials used (less than $2.5\%$ strain at operating pressure for low-density polyethylene \cite{LDPE}) we assume these circles maintain a constant diameter over the range of inflation pressure. This circular shape imposes the constraint that the tube diameter is the distance between two points $\vec{r}_o(s)$ and $\vec{r}_i(s)$, which in terms of the parameters can be expressed:
\begin{equation}
D=R_o-R_i.
\label{eqn:diam_constraint}
\end{equation}

For the final constraint, we make the assumption that the angle between the inner and outer tangent vectors remains constant at $2\theta$, where $\theta$ is the drawn angle of the actuator before actuation; this is based on observations of prototyped actuation and measurements of the prototyped helices pitch and radii. This offset angle will be a rotation about the shared normal vector. Because the inner and outer curves are separated by a constant distance, the curves form a Bertrand curve pair. This means the normal vectors of the curves will be aligned with each other, in the direction $(\vec{r}_o(s)-\vec{r}_i(s)),$ 
and the tangent vectors ($\vec{T}_i(s)$ and $\vec{T}_o(s)$) will be offset by a rotation about the shared normal \cite{hsiung1981first}. 
This relationship is diagrammed in Fig.~\ref{fig:GeoCon}. With this definition, the helix radii, particularly the inner radius $R_i$, can be positive or negative. In the case of a negative $R_i$, the normal vector of the Frenet-Serret frame points away from the central axis of the helix, as seen in Fig.~\ref{fig:ch5_Params}(b).

The normal and tangent vectors obtained from a Frenet-Serret frame of the parameterization in Equation~(\ref{eqn:Helix}) are:
\begin{equation}
\vec{T}(s)=\frac{\dot{\vec{r}}(s)}{ \|\dot{\vec{r}}(s)\|}=\frac{1}{\sqrt[]{b^2+R^2}}\begin{bmatrix} R\cos(s)\\ -R\sin(s)\\ b \end{bmatrix}
\label{eqn:tangent}
\end{equation}
\begin{equation}
\vec{N}(s) =\frac{\dot{\vec{T}}(s)}{ \|\dot{\vec{T}}(s)\|}= \begin{bmatrix} -\sin(s)\\ -\cos(s)\\ 0 \end{bmatrix}
\label{eqn:normal}
\end{equation}
where $\vec{T}(s)$ is the unit tangent vector and $\vec{N}(s)$ is the unit normal vector to the curve. These frames can be seen in Fig.~\ref{fig:ch5_Params}(b). The constant angle between the tangents can be captured with the dot product of the unit tangent vectors:
\begin{equation}
\vec{T}_i^T \vec{T}_o= \|\vec{T}_i\|\|\vec{T}_o\| \cos (2\theta)=\cos (2\theta)
\label{eqn:tangent_dot}
\end{equation}
Substituting in the tangent vectors $\vec{T}_i$ and $\vec{T}_o$ from Equation~(\ref{eqn:tangent}) into Equation~(\ref{eqn:tangent_dot}) we get the final constraint:
\begin{equation}
\cos(2\theta)=\frac{R_i R_o + b^2}{\sqrt[]{(b^2+R_i^2)(b^2+R_o^2)}}
\label{eqn:angle_constraint}
\end{equation}

\subsection{Closed-Form Uniform Actuation Solution}
\label{subsec:ch5_UniformActModel}
The three geometric constraints in Equations~(\ref{eqn:act_len_constraint}), (\ref{eqn:diam_constraint}), and (\ref{eqn:angle_constraint}) define the relationship between the actuator parameterization and the robot shape parameterization. In fact, the equations in their current form give a solution for the inverse problem, taking the desired robot helical shape (given by $R_o$, $R_i$, and $b$) and giving the actuator shape  to achieve it (defined by $D$, $\lambda$, and $\theta$). A forward solution can be solved from the equations as:

\begin{equation}
\begin{aligned}
R_o=\frac{D(1-\lambda\cos(2\theta))}{\lambda^2-2\lambda\cos(2\theta)+1}
\\
R_i=\frac{D\lambda(\cos(2\theta)-\lambda)}{\lambda^2-2\lambda\cos(2\theta)+1}
\\
b=\frac{D\lambda\sin(2\theta)}{\lambda^2-2\lambda\cos(2\theta)+1}
\end{aligned}
\label{eqn:Soln_Rb1}
\end{equation}
The radius of the center path of the soft robot body can be calculated using Equation~(\ref{eqn:ch5_centerPath}) and taking the average of the inner and outer radii solutions in Equation~(\ref{eqn:Soln_Rb1}):
\begin{equation}
    R_c=\frac{D(1-\lambda^2)}{2(\lambda^2-2\lambda\cos(2\theta)+1)}
    \label{eqn:Rc}
\end{equation}
With these equations we can calculate the resulting helical shape from a helical actuator configuration.

\begin{table*}[t!]
\caption{Summary of the forward and inverse static solutions of helical actuation}
\label{table:solution}
\begin{center}
\resizebox{\textwidth}{!}{
\begin{tabular} { c|c|c }
Inverse Solution & Radius and Pitch Parameterized & Curvature and Torsion Parameterized\\
& Forward Solution & Forward Solution\\
& &\\
\hline
& &\\
\vspace{-0.2cm}
$D=R_o-R_i$ & $R_i=\frac{D\lambda(\cos(2\theta)-\lambda)}{\lambda^2-2\lambda\cos(2\theta)+1}$ & $\kappa_i=\frac{\cos(2\theta)-\lambda}{D\lambda}$ \\
& &\\
\vspace{-0.2cm}
$\lambda=\frac{\sqrt[]{b^2+ R_i^2}}{\sqrt[]{b^2+ R_o^2}}$  & $R_c=\frac{D(1-\lambda^2)}{2(\lambda^2-2\lambda\cos(2\theta)+1)}$  & $\kappa_o=\frac{1-\lambda\cos(2\theta)}{D}$\\
& & \\
\vspace{-0.2cm}
$\theta=\frac{1}{2}\arccos\left(\frac{R_i R_o + b^2}{\sqrt[]{(b^2+R_i^2)(b^2+R_o^2)}}\right)$ & $R_o=\frac{D(1-\lambda\cos(2\theta))}{\lambda^2-2\lambda\cos(2\theta)+1}$  & $\tau_i=\frac{\sin(2\theta)}{D\lambda}$ \\ 
& &\\
\vspace{-0.2cm}
& $b=\frac{D\lambda\sin(2\theta)}{\lambda^2-2\lambda\cos(2\theta)+1}$ & $\tau_o=\frac{\lambda\sin(2\theta)}{D}$  \\
& &\\
\hline
\end{tabular}}
\end{center}
\end{table*}

These equations can be reformulated in terms of curvature and torsion to give additional insight, using the following substitution:
\begin{equation}
\kappa=\frac{R}{b^2+R^2}  \hspace{1cm}    \tau=\frac{b}{b^2+R^2}\\
\label{eqn:Rb_to_kT}
\end{equation}
where $\kappa$ is the helix curvature and $\tau$ is the helix torsion in the Frenet-Serret sense, parameterized by the actuated path length. With this substitution, there are now four parameters describing the robot body shape ($\kappa_o$, $\kappa_i$, $\tau_o$, and $\tau_i$) but still only three equations. The inner and outer pitch being equal adds a fourth constraint equation:
\begin{equation}
\frac{\tau_o}{\tau_o^2+\kappa_o^2}=\frac{\tau_i}{\tau_i^2+\kappa_i^2}
\end{equation}
Substituting the solutions from Equation~(\ref{eqn:Soln_Rb1}) into the variable re-parameterization in Equation~(\ref{eqn:Rb_to_kT}) gives:
\begin{equation}
\begin{aligned}
\kappa_o&=\frac{1-\lambda\cos(2\theta)}{D} \hspace{1cm} &\tau_o&=\frac{\lambda\sin(2\theta)}{D} \\ \\
\kappa_i&=\frac{\cos(2\theta)-\lambda}{D\lambda} &\tau_i&=\frac{\sin(2\theta)}{D\lambda}
\end{aligned}
\end{equation}
It should be noted that $\kappa_o$ will always be positive, but $\kappa_i$ will be negative if $\lambda>\cos(2\theta)$. This corresponds to situations where $R_i$ is also negative, which happens as $\lambda$ approaches 1 because the two diametrically opposed lines are on opposite sides of the helical center.

The geometric model can also be defined in terms of the kinematic variables used in Cosserat-rod theory, {$\vec{u}(t)$} and {$\vec{v}(t)$}, which are the linear and angular rates of change of a material-attached reference frame with respect to length along the undeformed material centerline (usually these frames are assigned such that they exhibit no torsion when the backbone is in the undeformed state.) The linear rate of change, {$\vec{v}(t)$}, includes the shear and the axial length change. While there is no shear, the axial component of $\vec{v}(t)$, $v_z$ is equal to the centerline compression ratio, $\lambda_c$:
\begin{equation}
    v_z=\lambda_c = \sqrt{\frac{\lambda^2+2\lambda\cos(2\theta)+1}{2(1+\cos(2\theta))}}
\end{equation}
which is derived by using the helical actuation solution in Equations~\ref{eqn:Soln_Rb1} and \ref{eqn:Rc} and comparing the initial length of the centerline to the actuated length of that path.

The angular rate vector (corresponding to material bending and torsion), $\vec{u}(t)=[u_x\ u_y\ u_z]^\top$, is related to the Frenet-Serret curvature and torsion of the center path. The magnitude of {$[u_x\ u_y]^\top$} can be calculated using Equation~(\ref{eqn:Rc}), with a scaling correction of $\lambda_c$ to correct for the fact that the Cosserat derivatives are taken with respect to length along the uncompressed centerline, while the geometric curvature is a rate of change with respect to length along the compressed centerline: 
\begin{equation}
\sqrt{u_x^2+u_y^2}=\frac{2(1-\lambda^2)}{D(\lambda^2+2\lambda\cos(2\theta)+1)}\lambda_c
\end{equation}
The scaled Frenet-Serret torsion using Equation~(\ref{eqn:Rc}) and the scaling correction:
\begin{equation}
\tau_c=\frac{4\lambda\sin(2\theta)}{D(\lambda^2+2\lambda\cos(2\theta)+1)}\lambda_c
\end{equation}
represents the torsion of a material frame which was assigned such that one axis always intersects the tendon line. To calculate the torsion $u_z$ of a more conventional material frame which was assigned with zero torsion in the undeformed ($\lambda=1$) state, we can subtract out the reference torsion as follows:
\begin{equation}
    u_z=\tau_c-\frac{2\sin(2\theta)}{D(1+\cos(2\theta))}
\end{equation}

While the remainder of this paper only uses the geometric model in Equations~(\ref{eqn:Soln_Rb1}) and (\ref{eqn:Rc}), these additional parameterizations could be used in the future to compare the results from the geometric model to other common parameterizations of lines in 3D and to Cosserat-rod-based models for continuum robots.
The forward and inverse solutions for the helical actuation are summarized in Table~\ref{table:solution}.

\subsection{General Actuation}
\label{subsec:ch5_GenActModel}
\begin{figure}[tb!]
\centering
	\includegraphics[width=\columnwidth]{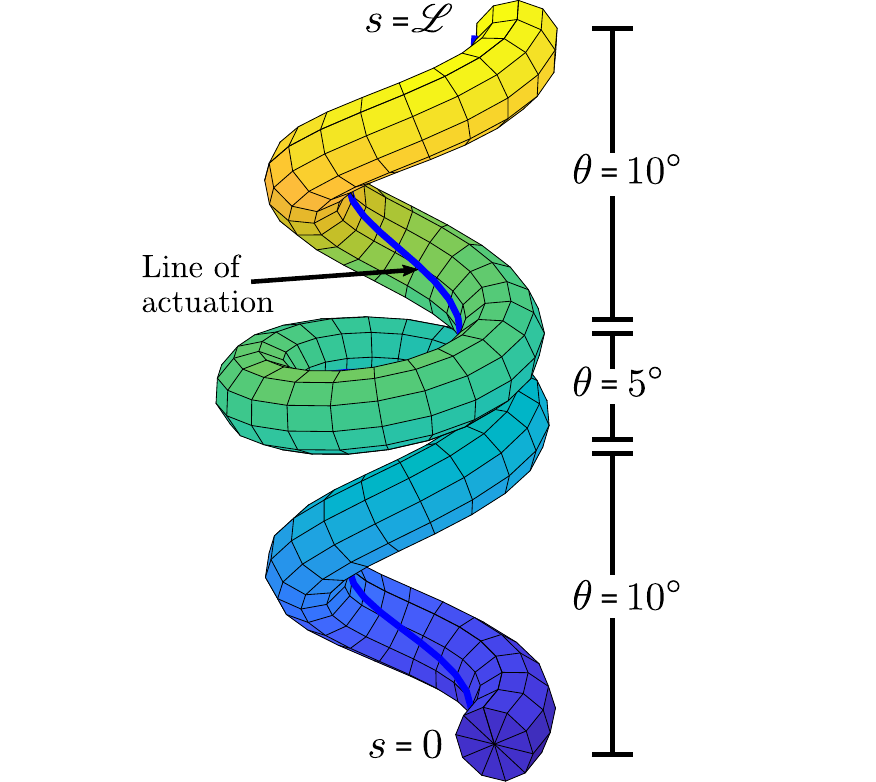}
	\caption{\small Example shape generated by a tendon routing with changing angle. Actuator starts with $\theta=10^{\degree}$, transitions in the center to $\theta=5^{\degree}$, and ends back at $\theta=10^{\degree}$.}
	\label{fig:ch5_MultiHelices}
\end{figure}
The geometric model generated above covers a larger range of actuation than 2D constant curvature models alone, but still requires the actuation to be uniform along the length. A truly general model would allow change of the actuator parameters, $\theta$ and $\lambda$, along the line of actuation. To do this, we take further inspiration from constant curvature models. For multi-segment continuum manipulators, constant curvature models can be applied consecutively to each segment of the continuum robot, creating a piecewise constant curvature shape. Similarly, the constant helical model can be applied in a piecewise manner to a piecewise helical shape (Fig.~\ref{fig:ch5_MultiHelices}). Whereas constant curvature models generally apply this idea to serial chains of independently actuated continuum modules, we can apply the same idea while imagining varying the actuator angle, actuator contractio, or both of a single actuator, depending on the actuator used. Fig.~\ref{fig:ch5_MultiHelices} shows an example where the tendon is routed along the tube in three uniform helical sections, routed with $\theta$ equal to $10^{\degree}$, then $5^{\degree}$, and then back to $10^{\degree}$, with the value of $\lambda$ held constant. While each section is a uniform helix, together they produce a non-uniform shape that, through growing, can be deployed sequentially from base to tip.

\begin{figure}[tb!]
\centering
	\includegraphics[width=\columnwidth]{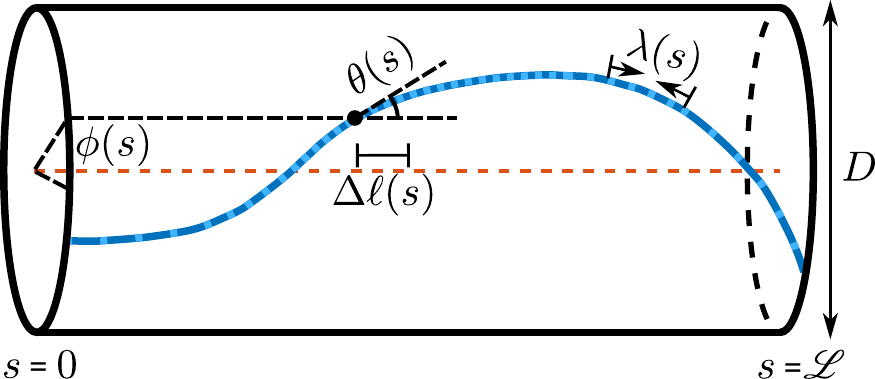}
	\caption[Parameterization of a general tendon routing (blue).]{\small Parameterization of a general tendon routing (blue). The centerline of the tube is shown in red. All parameters are a function of the length along the centerline, $s\in [0,\mathscr{L}]$. The helix parameters, $\theta(s)$ and $\lambda(s)$, change along the length of the actuator, while the diameter, $D$, is constant. Two new parameters are shown, the angle of the actuator around the tube, $\phi(s)$, and the discretized length of tube each constant segment is applied for, $\Delta\ell(s)$.}
	\label{fig:ch5_GenActSch}
\end{figure}

By changing the actuator parameters over smaller and smaller sections, we get a general actuator shape that we can calculate as a series of uniform helical sections (Fig.~\ref{fig:ch5_GenActSch}). This general actuator requires a slightly more complex parameterization, with $\theta$ and $\lambda$ now functions of the position along the tube, $s\in [0,\mathscr{L}]$, where $\mathscr{L}$ is the total length of the tube. The parameter $s$ used here is different from that used in the above sections and describes the length along the centerline of the tube, not along the arc length of the actuator. While it is also theoretically possible to have the diameter, $D$, vary along the tube, this is difficult to achieve in a physical system. We introduce an additional parameter for the continuous general actuator, $\phi(s)$. This new parameter is the angle of the actuator around the tube from a fixed point on the tube. It is a function of the other parameters as:
\begin{equation}
    \phi(s)=\frac{2}{D}\int_{t=0}^s \tan(\theta(t))dt
    \label{eqn:ch5_phi(s)}
\end{equation}
Because this is an integral of $\theta(s)$, $\phi(s)$ describes the current position of the actuator relative to the tube at each length, as opposed to $\theta(s)$, which describes the rate of change in the actuator position. This makes $\phi(s)$ useful for plotting and implementing actuator designs. 

While a truly continuous actuator path can be described using the parameters above, for practical implementation, both in the model and in the actuator construction, we discretize these parameters. To do this, we discretize $s$ for an actuator path with $N$ helical sections as $s_j$ for $j\in[0,N]$, where $s_0=0$ and $s_{N}=\mathscr{L}$. Each segment, starting at $s_j$ for $j\in[0,N-1]$, has actuator parameters $\lambda$ and $\theta$ associated with it and, for the discrete formulation, we introduce the parameter $\Delta\ell(s_j)=s_{j+1}-s_j$, which represents the length of tube that a particular uniform actuation is applied for. This parameter is needed since the discretization does not need to be uniform in $s$. In the limit where actuator parameters change continuously, $\Delta\ell(s_j)$ is the differential, $ds$. This also changes the calculation of $\phi(s)$ in Equation~(\ref{eqn:ch5_phi(s)}) slightly to be a sum instead of an integral:
\begin{equation}
    \phi(s_j)=\frac{2}{D}\sum_{k=0}^j \tan(\theta(s_k))\Delta\ell(s_k).
    \label{eqn:ch5_phi(si)}
\end{equation}

\begin{figure*}[bt!]
\hspace{0.7cm} (a) Mechanically Programmed \hspace{2.0cm} (b) Tendon and Stopper \hspace{2.3cm} (c) Pneumatic Muscles
\begin{center}
	\includegraphics[width=2.0\columnwidth]{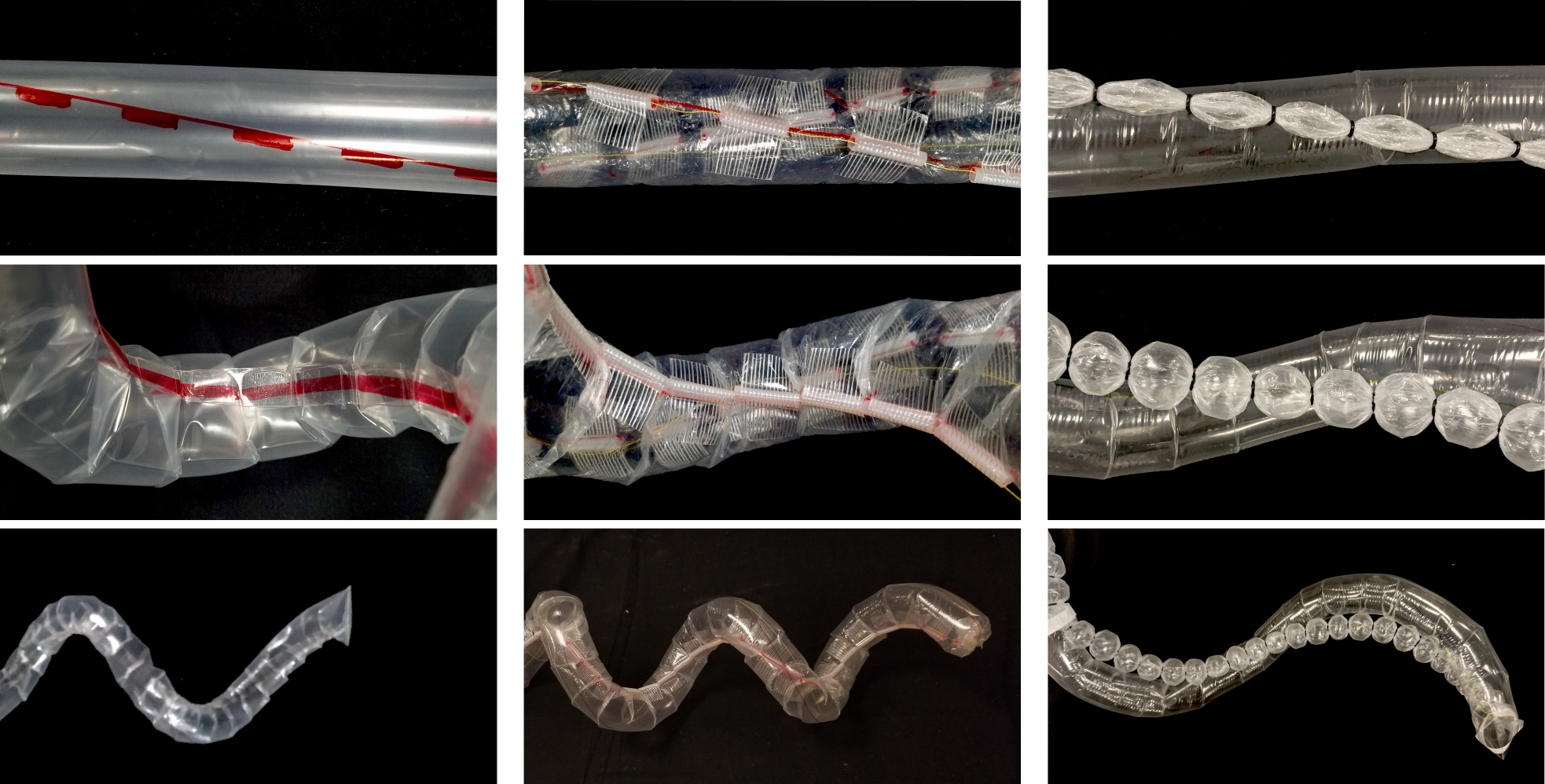}
\end{center}

	\caption[Implementation of general tendon actuation.]{\small Implementation of actuation. (a) Mechanically programmed implementation uses discrete removal of material using tape to achieve a single static shape that can be grown. (b) Tendon and stopper implementation creates one shape when the string is fully relaxed and another when the stoppers are fully connected, allowing actuation between two shapes. (c) Pneumatic artificial muscle implementation allows for approximately continuous change of $\lambda$ during actuation. The material wrinkles along the full line of actuation to reduce length.}
	\label{fig:HelixImplementation}
\end{figure*}

To implement this discretized formulation, we use a transformation matrix along a uniform helix which starts with the Frenet-Serret frame aligned to the transformation reference frame:
\begin{equation}
     \mathbf{T}^{j+1}_{j}(s_j)=\begin{bmatrix}\\&\mathbf{R}^{j+1}_{j}(s_j& & \vec{P}^{j+1}_{j}(s_j) \\ \\0 & 0 &0 & 1 \end{bmatrix}\\
     \label{eq:ch5_transformationCC}
\end{equation}
\begin{equation}
    \mathbf{R}^{j+1}_{j}(s_j)=\begin{bmatrix} \frac{R^2}{L^2}\cos{\frac{\Delta\ell_{\lambda}}{L}}+\frac{b^2}{L^2} & -\frac{R}{L}\sin{\frac{\Delta\ell_{\lambda}}{L}} & \frac{Rb}{L^2}(1-\cos{\frac{\Delta\ell_{\lambda}}{L}})\\ \\ \frac{R}{L}\sin{\frac{\Delta\ell_{\lambda}}{L}} & \cos{\frac{\Delta\ell_{\lambda}}{L}} & -\frac{b}{L}\sin{\frac{\Delta\ell_{\lambda}}{L}}\\ \\ \frac{Rb}{L^2}(1-\cos{\frac{\Delta\ell_{\lambda}}{L}}) & \frac{b}{L}\sin{\frac{\Delta\ell_{\lambda}}{L}} & \frac{b^2}{L^2}\cos{\frac{\Delta\ell_{\lambda}}{L}}+\frac{R^2}{L^2} \end{bmatrix}
    \label{eq:ch5_rotationCC}
\end{equation}
\begin{equation}
    \vec{P}^{j+1}_{j}(s_j)=\begin{bmatrix} \frac{R^2}{L}\sin{\frac{\Delta\ell_{\lambda}}{L}}+\frac{b^2}{L}\frac{\Delta\ell_{\lambda}}{L} \\ \\ R(1-\sin{\frac{\Delta\ell_{\lambda}}{L}}) \\ \\ \frac{Rb}{L}(\frac{\Delta\ell_{\lambda}}{L}-\sin{\frac{\Delta\ell_{\lambda}}{L}})\end{bmatrix}
    \label{eq:ch5_positionCC}
\end{equation}
where $\mathbf{T}^{j+1}_{j}(s_j)$ is the incremental transformation matrix along the length at segment $s_j$, $\mathbf{R}^{j+1}_{j}(s_j)$ is the rotation matrix between the frames of segments $s_{j}$ and $s_{j+1}$, $\vec{P}^{j+1}_{j}(s_j)$ is the position vector between segment $s_j$ and the next segment, $L=\sqrt{R^2+b^2}$, and $\Delta\ell_{\lambda}$ is the length of the path at that segment after the actuator deforms. This transformation is a function of the current segment, $s_j$, since $R$, $b$, $L$, and $\Delta\ell_{\lambda}$ depend on the actuator parameters at that segment. For most paths, we apply this transformation along the center axis, which means:
\begin{equation}
    \Delta\ell_{\lambda}=\Delta\ell(s_j)\lambda_c=\Delta\ell(s_j)\sqrt{\frac{\lambda^2+2\lambda\cos(2\theta)+1}{2(1+\cos(2\theta))}}.
\end{equation}
The transformation allows us to progressively add short helical sections and align the starting and ending frames between sections, so we can calculate the transformation from the path to the world frame, $\mathbf{T}^{j+1}_{w}$, recursively as:
\begin{equation}
    \mathbf{T}^{j+1}_{w}=\mathbf{T}^{j}_{w}\mathbf{T}^{j+1}_{j}(s_j)
    \label{eqn:ch5_iterativeTransform}
\end{equation}
where the position vector of the transformation matrix $\mathbf{T}^{j}_{w}$, $\vec{P}^{j}_{w}$, is equal to the $j$th point on the actuated shape, $\vec{r}_c(s_j)$, and $\mathbf{T}^{0}_{w}=I$, the identity matrix. For each section, we use the uniform helix model in Table~\ref{table:solution} to take the actuator parameters, $\theta$, $\lambda$, and $D$, of that actuator segment and calculate the shape, $R_c$ and $b$, which can then be used to calculate the transformation, $\mathbf{T}^{j+1}_{j}(s_j)$, using Equations~(\ref{eq:ch5_transformationCC})-(\ref{eq:ch5_positionCC}). To get the inner and outer paths of the shape, additional transformations are needed between the center line and the inner or outer lines:
\begin{equation}
     \mathbf{T}^c_{i}(s_j)=\begin{bmatrix} \cos(\alpha_{c,i}) & 0 & -\sin(\alpha_{c,i}) & 0 \\ 0 & 1 & 0 & \frac{D}{2} \\ \sin(\alpha_{c,i}) & 0 & \cos(\alpha_{c,i}) & 0\\0 & 0 &0 & 1 \end{bmatrix}\\
     \label{eq:ch5_transformationIC}
\end{equation}
\begin{equation}
     \mathbf{T}^c_{o}(s_j)=\begin{bmatrix} \cos(\alpha_{c,o}) & 0 & \sin(\alpha_{c,o}) & 0 \\ 0 & 1 & 0 & -\frac{D}{2} \\ -\sin(\alpha_{c,o}) & 0 & \cos(\alpha_{c,o}) & 0\\0 & 0 &0 & 1 \end{bmatrix}\\
     \label{eq:ch5_transformationOC}
\end{equation}
where the angles $\alpha_{c,i}$ and $\alpha_{c,o}$ are the angle difference between the center path tangent and the inner and outer path tangents, respectively, defined as:
\begin{equation}
\begin{aligned}
    \alpha_{c,i}=\arctan\left(\frac{b}{R_i}\right)-\arctan\left(\frac{b}{R_c}\right) \\
    \alpha_{c,o}=\arctan\left(\frac{b}{R_c}\right)-\arctan\left(\frac{b}{R_o}\right)
    \end{aligned}
\end{equation}
where $b$, $R_i$, $R_c$, and $R_o$ are defined in Table~\ref{table:solution}. We post multiply the centerline transformation matrix, $\mathbf{T}^{j}_{w}$, by these matrices to get the inner and outer paths in the world frame. 
These transformations move the points along the normal vector shared by inner, outer, and center paths, and rotate them about that normal vector. In order to implement the shape change described by these equations, we need actuation methods to shorten defined paths along the body, which will be discussed in the following section.

\section{Actuator Implementation}
\label{sec:ch5_actImplementation}
We implemented the actuation of our soft inflated robot body using three different methods, as shown in Fig.~\ref{fig:HelixImplementation}: (1) mechanically programmed shapes, (2) tendon actuation with stoppers, and (3) pneumatic artificial muscle actuation. Actuation methods are constrained by the compliance needed to evert in order to grow. These methods allow different amounts of control, from growth into a single set shape to actuation among a range of shapes. In the remaining sections, the soft inflated robot bodies are formed from low-density polyethylene (LDPE) \cite{LDPE}  in a range of diameters and material thickness, primarily supplied by Aviditi (Elgin, IL). This section has appeared previously in \cite{blumenschein2018helical} but has been revised and included here for completeness.

\subsection{Mechanically Programmed}
\label{subsubsec:mech_prog}
We refer to the first implementation as being “mechanically programmed,” meaning that the shape is permanently constructed during the manufacturing process. The robot body can then be grown into this predetermined shape. In tasks where the desired path is known or can be planned ahead of time, this implementation allows precise shape control and allows us to create static shapes. Despite this implementation not being ``actuated" in the sense that it changes shape actively, we can use the same models of general actuation to describe the permanent shapes achieved, which is useful for creating shapes to test the model.

The robot body is mechanically programmed by removing discrete sections of material along the actuator path. These sections are manually pinched together and held by tape. Though the actual description of the model describes a proportional shortening along the entire length of actuation, this can be well approximated by alternating pinched and straight sections at a sufficiently tight spacing, as seen in Fig.~\ref{fig:HelixImplementation}(a).

\subsection{Tendon and Stoppers}
\label{subsubsec:tendon_stopper}
The second implementation uses a combination of tendons and rigid stoppers to actuate between two shapes, usually a straight tube and a single desired shape. The tendon provides the force to the tip of the robot and the stoppers mechanically limit the contraction when they connect. This is done by arranging alternating gaps and stoppers along the line of actuation (Fig.~\ref{fig:HelixImplementation}(b)). A tendon is fed through the stoppers and attached to the far end of the robot. When this string is pulled, the gaps collapse along the line of actuation and only the stoppers are left. This is similar to the mechanically programmed implementation, in which discrete sections are fully wrinkled and the remaining material is left extended. In this implementation, the model can only be used when all the stoppers are connected. The value of $\lambda$ will be the ratio between the stopper length and the total length for a section.

We created this actuation using PTFE tubing for the stoppers and high molecular weight polyethylene braided line for the tendon. This combination provided a low coefficient of friction, which is beneficial since the force needed to pull the cable increases as more curves are formed in the path \cite{kaneko1991basic}. The tendon implementation can only be used to accurately actuate between two discrete shapes because a specified value for $\lambda$ along the line of actuation will not be guaranteed until the cable is fully actuated and all the stoppers are touching. In practice, this happens because, as the cable is actuated, the tube wall will buckle and the tube will bend first at a single point. This buckled point will have a much lower stiffness than the rest of the tube and will require only a small force for each additional displacement in order to change the volume. This point will continue to be the location of bending until either that small force is greater than the force to produce a new buckle or until the stoppers touch. When either of these occurs, a new buckling point will appear. This will repeat until all the stoppers are touching, at which point the specified actuation can be guaranteed. While the tendon actuation shown here is achieved by pulling on the tendon by hand, this actuation method has been previously demonstrated using DC motors to autonomously shape helical tendon paths \cite{gan20203d}.

\subsection{Pneumatic Artificial Muscles}
\label{subsec:PAMs}
Pneumatic muscles are a class of actuators that change length or shape based on the internal pressure in the actuator \cite{chou1996measurement,Hawkes2016,Greer2017}. When made uniformly, the muscles will have uniform contraction or expansion along the length. 

For our final implementation we use a type of pneumatic artificial muscles (PAMs) called series pneumatic artificial muscles (sPAMS) \cite{Greer2017}. sPAMs contract with increasing internal pressure and can be constructed of the same inextensible plastic used to construct the robot bodies. By constructing a robot with a pneumatic artificial muscle along the line of actuation, we can shorten the full line of actuation simultaneously, and the actuation will be approximately uniform given a sufficiently short robot (less than 1 meter long) \cite{coad2019vine}. The use of sPAMs on growing robots was first shown in \cite{greer2019soft}, which demonstrated that growth and steering using sPAMs act as independent degrees of freedom. While previous implementations used tabs heat-sealed to the body to attach the sPAM, in order to replicate a general line of actuation, an sPAM actuator can be attached along the desired line of actuation with a soft viscoelastic adhesive (TrueTape, LLC). For the purposes of the model, we assume that the contraction ratio, $\lambda(s)$, will be the same along the entire length of the actuator at any given time, and that the contraction is a function of the pressure within the actuator and the pressure within the robot \cite{Greer2017}. Therefore, while the value of $\lambda(s)$ will be the same at all points along the robot, we can continuously change the $\lambda$ value within a range over time by changing the pressure, while maintaining a set $\theta$ and $D$ (Fig.~\ref{fig:HelixImplementation}(c)). Initial validation of this assumption was carried out with a $0.5$~m long robot with $\theta$ varying between $5^{\degree}$ and $10^{\degree}$ along the length. As the actuator pressure was changed, measurements showed an average error in the expected contraction ratio of $2.1\%$ and a maximum error of $3.8\%$, with no clear relationship with length or pressure and only a small increase with varying angle ($2.4\%$ versus $1.7\%$). Unlike the mechanically programmed and tendon and stopper implementations, the value of $\lambda$ is not inherently known from the construction, so either a mapping must be developed to relate the pressure to the shape or another measurement of the actuator strain will be needed.

\section{Model Validation}
\label{sec:ch5_modelValidation}
We experimentally validate the geometric models introduced in Section~\ref{sec:ch5_geomModel}, using the implementations described in Section~\ref{sec:ch5_actImplementation}. We start by showing validation of uniform helical actuation for both static and active shapes, and then move on to experiments on generally routed actuators, again static and active.
\subsection{Uniform Actuation}
To validate the geometric model for uniform helical actuation, we built both static and active shapes with uniform actuation along the tube length. After extracting the 3D path of the resulting shapes, we ran a global optimization to find the best match between the data, $r(s)$, and the vertically aligned helix parameterization in Equation~(\ref{eqn:Helix}). In all situations, we optimized the initial orientation, $\mathbf{R}_0$, and initial position, $P_0$, of the data:
\begin{equation}
    \mathbf{R}_0=\mathbf{R}_x(\gamma_1)\mathbf{R}_z(\gamma_2)\mathbf{R}_x(\gamma_3)
    \label{eq:ch5_rotation0}
\end{equation}
\begin{equation}
    P_{0}=[x_0, y_0, z_0]^T
    \label{eq:ch5_position0}
\end{equation}
where $\mathbf{R}_x$ and $\mathbf{R}_z$ are rotations about the $x$ and $z$ axis, respectively, $\gamma_1$, $\gamma_2$, and $\gamma_3$ are Euler angles, and $x_0$, $y_0$, and  $z_0$ are the offsets in $x$, $y$, and $z$ directions respectively. The data was rotated first, $r_{rot}(s)=\mathbf{R}_0r(s)=[x, y, z]^T$. Then the $x$ and $y$ components of the rotated data were compared to the helix equations in Equation~(\ref{eqn:Helix}), rewritten as a function of $z$ and including the initial position:
\begin{equation}
\begin{aligned}
    x_m(s) = R\cos\left(\frac{z(s)}{b}+z_0\right)+x_0\\
    y_m(s) = R\sin\left(\frac{z(s)}{b}+z_0\right)+y_0
\end{aligned}
\end{equation}
with the optimization minimizing the error function:
\begin{equation}
    e = \sum_s(x(s)-x_m(s))^2+(y(s)-y_m(s))^2.
    \label{eq:error_func}
\end{equation}

\subsubsection{Static Helices}
\begin{figure}[t!]
\centering
	\includegraphics[width=\columnwidth]{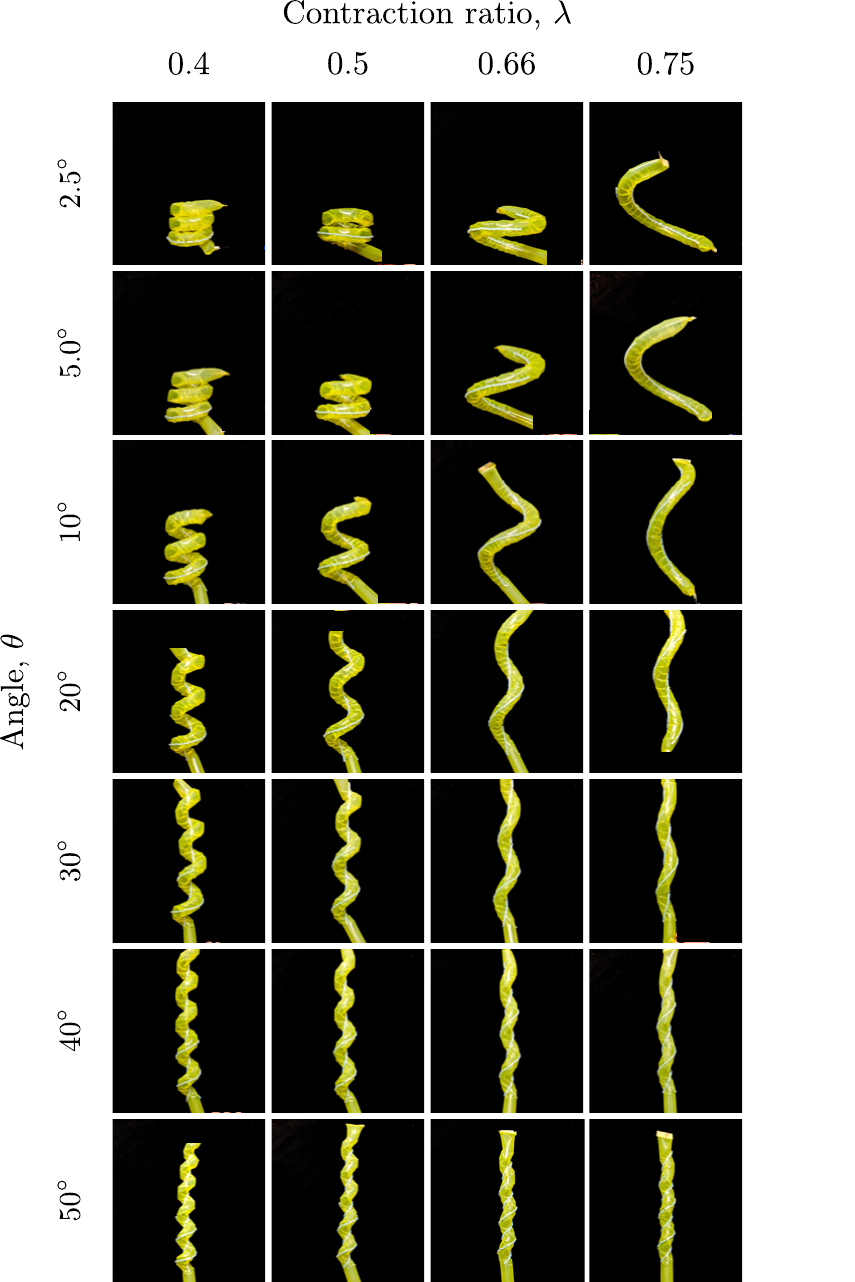}
	\caption[Static helix prototypes for a range of actuator parameters, $\theta$ and $\lambda$, with a tube diameter, D, of 2.62~cm.]{\small Static helix prototypes for a range of actuator parameters, $\theta$ and $\lambda$, with a tube diameter, D, of 2.62~cm. The shapes are unactuated and produced using the mechanically programmed implementation (Fig.~\ref{fig:HelixImplementation}(a)). The length of the inner path was 30.5~cm for each shape.}
	\label{fig:ch5_StaticHelices}
\end{figure}

To test a large range of the parameters, we built static shapes for $\lambda\in[0.4,0.75]$ and $\theta\in[2.5^{\degree},50^{\degree}]$, with a tube diameter of 2.62~cm and material thickness of $60~\mu$m (Fig.~\ref{fig:ch5_StaticHelices}). The length of the actuated path, the inner helical path, was 30.5~cm for all shapes.  Qualitatively, changes in $\theta$ changed the relationship between the pitch and radius of the path, modifying the slope of the helix, while changes in $\lambda$ tightened or loosened the helix.

\begin{figure*}[t!]
\centering
	\includegraphics[width=2.0\columnwidth]{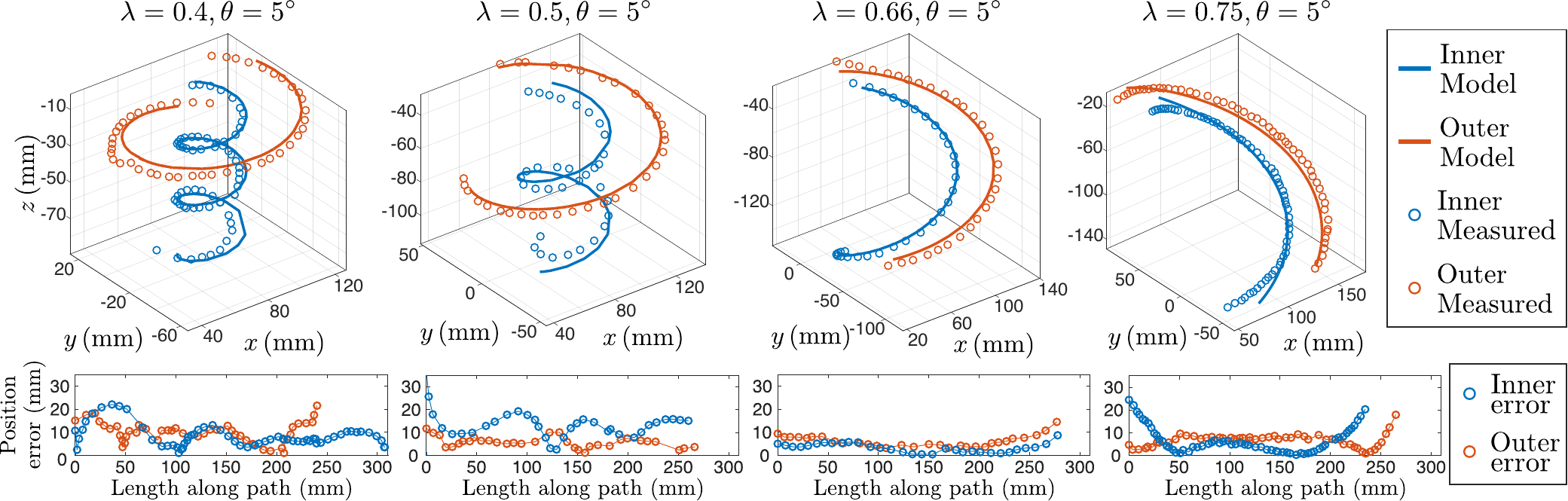}
	\caption{\small Examples of matching between the constructed shape and the modeled helix for $\theta=5^{\degree}$ and $\lambda=[0.4,0.5,0.66,0.75]$. Measurements are of the static helices shown in Fig.~\ref{fig:ch5_StaticHelices}. The inner and outer paths were matched simultaneously using the same initial orientation and position. Top row shows 3D paths of the model and measured data while bottom row shows the position error for inner and outer paths.}
	\label{fig:ch5_StaticHelixExp}
\end{figure*}

The shapes were measured using a magnetic tracker system (Ascension Technology Corporation trakSTAR). The tracking system measured the position and orientation of a sensor probe relative to a stationary transmitter, with a spatial accuracy of 1.4~mm RMS and an update rate of $80$~Hz. The measurements were read and saved into a text file using a C++ program. The sensor probe was $2.0$~mm in diameter and was attached to a cable $30.5$~cm long. Measurements were made by running the sensor probe along the inner or outer path of the shape. To achieve this without applying forces that would distort the shape, after the shape was constructed we taped short, rigid tubes along the path to be measured, pushed the sensor probe and cable through the tubes up to a length of $30.5$~cm, and then slowly pulled the probe through the tubes by hand while measuring its position relative to the transmitter.  The probe was light enough to avoid deforming the shape, but the shape could still be temporarily distorted while the probe was moved, so we only recorded data points with the probe stationary, and we took the average value over 1-2~s of measurements to lessen the effects of noise. The prototypes were mounted to a surface throughout the measurements so that both the inner and outer helix paths could be measured with the prototype shape in the same orientation. These data sets were aligned vertically and matched with the model according to the optimization described above. The helix parameters, $R_i$, $R_o$, and $b$, were calculated from the known actuator design parameters, and the error between the paths was minimized over $\gamma_1$, $\gamma_2$, $\gamma_3$, $x_0$, $y_0$, and $z_0$. Both inner and outer paths were fit simultaneously, and the error of the paths (Equation~(\ref{eq:error_func})) were summed to create a combined error. Examples of the resulting match between modeled and measured shapes, including the position error as a function of length along the path, are shown in Fig.~\ref{fig:ch5_StaticHelixExp}. The alignment of the shapes was most successful when more revolutions of the helical path were measured, so paths with high values of $\theta$ and low values of $\lambda$ were easier to orient.

\begin{figure}[t!]
\centering
	\includegraphics[width=\columnwidth]{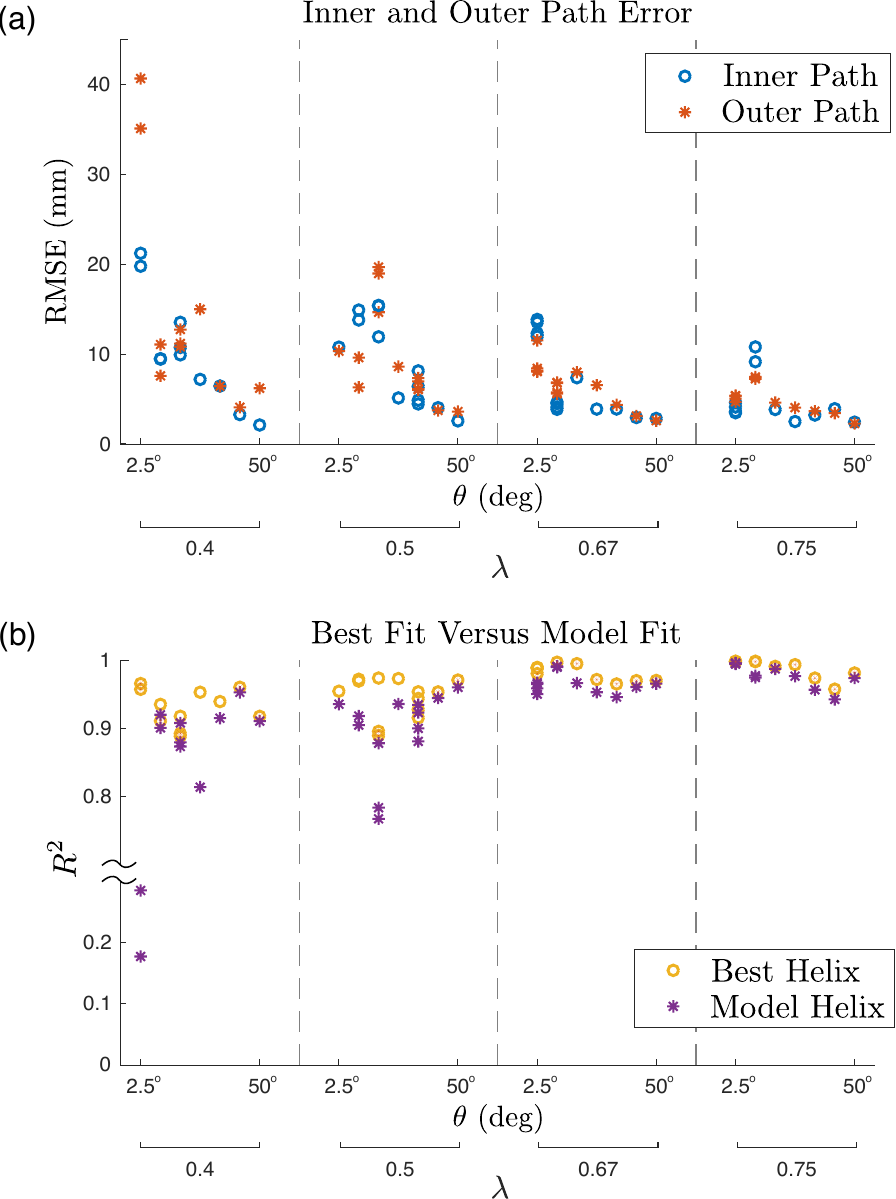}
	\caption[Metrics quantifying the fit of the model to the measured shape for the test helices.]{\small Metrics quantifying the fit of the model to the measured shape for the static test helices in Fig.~\ref{fig:ch5_StaticHelices}. (a) The root mean square error (RMSE) of the innermost and outermost helices. Errors are low on average relative to the length of the robot measured and the largest RMSE can be explained by self-collision preventing the tube from reaching the predicted shape. (b) The $R^2$ value of the model predicted helix compared to the $R^2$ for fitting the best helix for the data. For the majority of shapes, the model provides nearly as good a fit as the best helix.}
	\label{fig:ch5_StaticHelixMeas}
	\vspace{-0.2cm}
\end{figure}

We measured and fit all 28 produced shapes. To quantify the fit, we used the root mean square error (RMSE) of the data to the helical path, calculated separately for the inner and outer paths, and the $R^2$ value to quantify both fits together (Fig.~\ref{fig:ch5_StaticHelixMeas}). Shapes with higher RMSE during initial fitting were measured and fit multiple times to check consistency of the measurement technique and help minimize human error. In Fig.~\ref{fig:ch5_StaticHelixMeas}(a), we can see that the RMSE error overall is relatively small, less than 15~mm over a length of 300~mm, with the exception of the $\theta =2.5^{\degree}$, $\lambda=0.4$ point. This much higher error can be explained by looking at the predicted pitch from the model, $b=0.252$~cm, which give a rise for one revolution of $2\pi b = 1.58$~cm, less than the tube diameter, 2.62~cm. This self-collision prevents the helix from taking on the predicted shape. In general, self-collision in helical actuation, which is not considered in the model, occurs when at least one revolution is completed and $2\pi b<D$ or equivalently, using the equation for $b$ (Equation~(\ref{eqn:Soln_Rb1})), when $\pi\sin(2\theta)+\cos(2\theta)<\frac{\lambda^2+1}{2\lambda}$. Predicting similar self-collision for general actuator routings is a topic of future study. Looking at the remaining shapes, there appears to be a trend where higher angles and smaller strains, i.e.\ larger values of $\lambda$, lead to less error. This is likely explained by larger $\theta$ making shapes with tighter profiles less easily deformed by external forces like gravity, while larger $\lambda$ leads to fewer revolutions for the same actuator length, giving less room for error in construction to affect the match to the model. In the model helix $R^2$ data in Fig.~\ref{fig:ch5_StaticHelixMeas}(b), we can similarly see that larger values for the contraction ratios, $\lambda$, have better fit (higher $R^2$) on average. 

To further examine the model, we compared the results of fitting the model to fitting the best helices for both inner and outer path. The same optimization and error function apply (Equations (\ref{eq:ch5_rotation0})-(\ref{eq:error_func})), but $R_i$, $R_o$, and $b$ are added to the list of variables being fit, instead of coming from the model. The only restriction here is that the inner and outer paths must share the same pitch, since this is required for the two helices to be attached to the same tube. The $R^2$ value of this fit is added to the plot in Fig.~\ref{fig:ch5_StaticHelixMeas}(b) as the best helix fit. The difference between the model and best helix fits indicates what amount of error lies in the model, and what amount can be attributed to measurement or implementation error. We can see that the difference in the $R^2$ values is small, again with the exception of the small $\theta$, small $\lambda$ shape. Not counting this self-interfering shape, fitting the best helix possible leads to a small increase in the average $R^2$ value from 0.94 to 0.96.

\begin{figure}[t!]
\centering
	\includegraphics[width=\columnwidth]{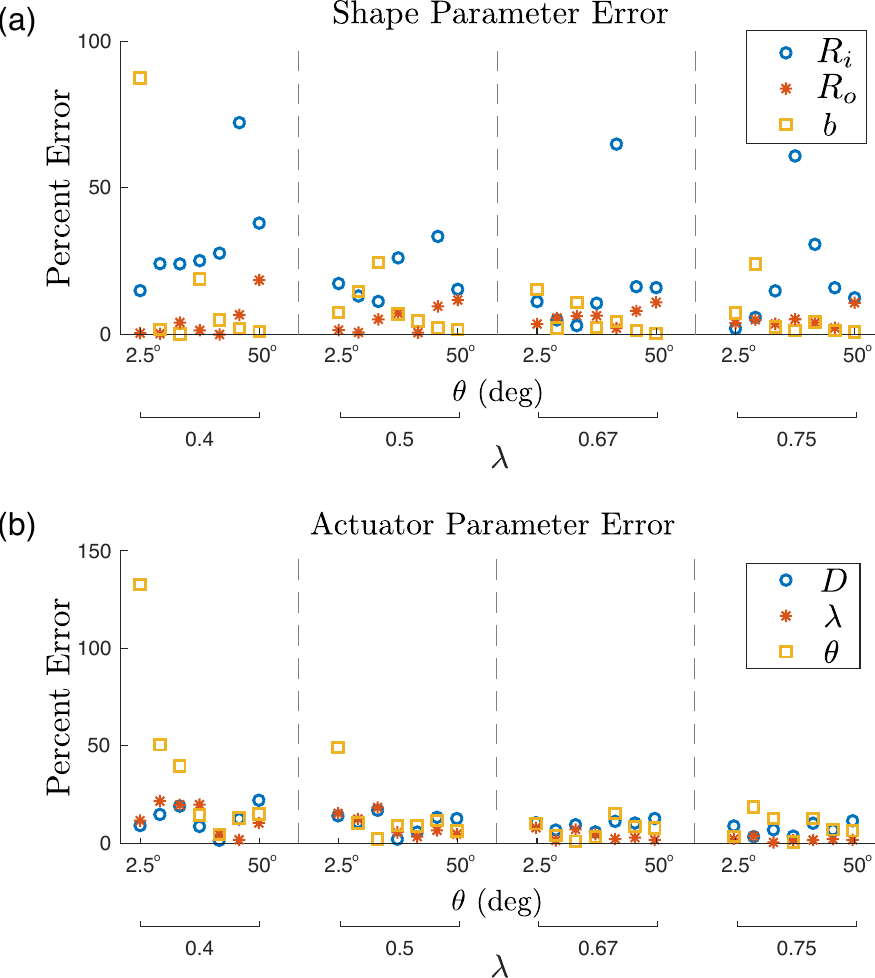}
	\caption[Errors in the shape and actuator parameters for the best fit helix compared to the model helix.]{\small Errors in the shape and actuator parameters for the static helices in Fig.~\ref{fig:ch5_StaticHelices}. These errors compare the best fit helix parameters (Fig.~\ref{fig:ch5_StaticHelixMeas}(b)) to the model helix parameters and show the sensitivity of the helix shape to changes in the parameters. (a) Percent error of the helix shape parameters shows low error in the pitch parameter and outer radius, and higher error in the inner radius. (b) The actuator parameters show low error overall, with the highest error in a few shapes occurring in the angle of the actuator $\theta$.}
	\label{fig:ch5_ParameterError}
\end{figure}

Finally, we looked at the difference in the shape and actuator parameters between the model helix and the best fit helix (Fig.~\ref{fig:ch5_ParameterError}). We calculated the percent difference of the best helix fit parameters compared to the model helix parameters. In Fig.~\ref{fig:ch5_ParameterError}(a), we can see the error in $R_i$, $R_o$, and $b$. The parameter error, combined with the data in Fig.~\ref{fig:ch5_StaticHelixMeas}(b), indicates where the helix shape is most and least sensitive to changes in the parameters. With a few exceptions, the percent error is below 15\% for most parameters. This error is largest for $R_i$ because the value is near zero for many of the shapes tested. As expected, the self-interfering shape registers a high error in the pitch parameter. Using the inverse model equations in Table~\ref{table:solution}, we calculate the actuator parameters for the best fit helix. The error in these actuator parameters compared to the true parameters is shown in Fig.~\ref{fig:ch5_ParameterError}(b). The largest errors are generated in $\theta$ for the shapes with the lowest value of $\theta$.

It should be noted that we focused on the variations from changing $\theta$ and $\lambda$ in our validation of the model. Diameter variation was not explored as deeply because the equations for $R_o$, $R_i$, and $b$ are all proportional to $D$ as seen in Table~\ref{table:solution}. So the expected effect of changing $D$ is a direct scaling of the resulting shape, given the same $\lambda$ and $\theta$. This was verified in~\cite{blumenschein2018helical}.

\begin{figure}[b!]
\centering
	\includegraphics[width=0.95\columnwidth]{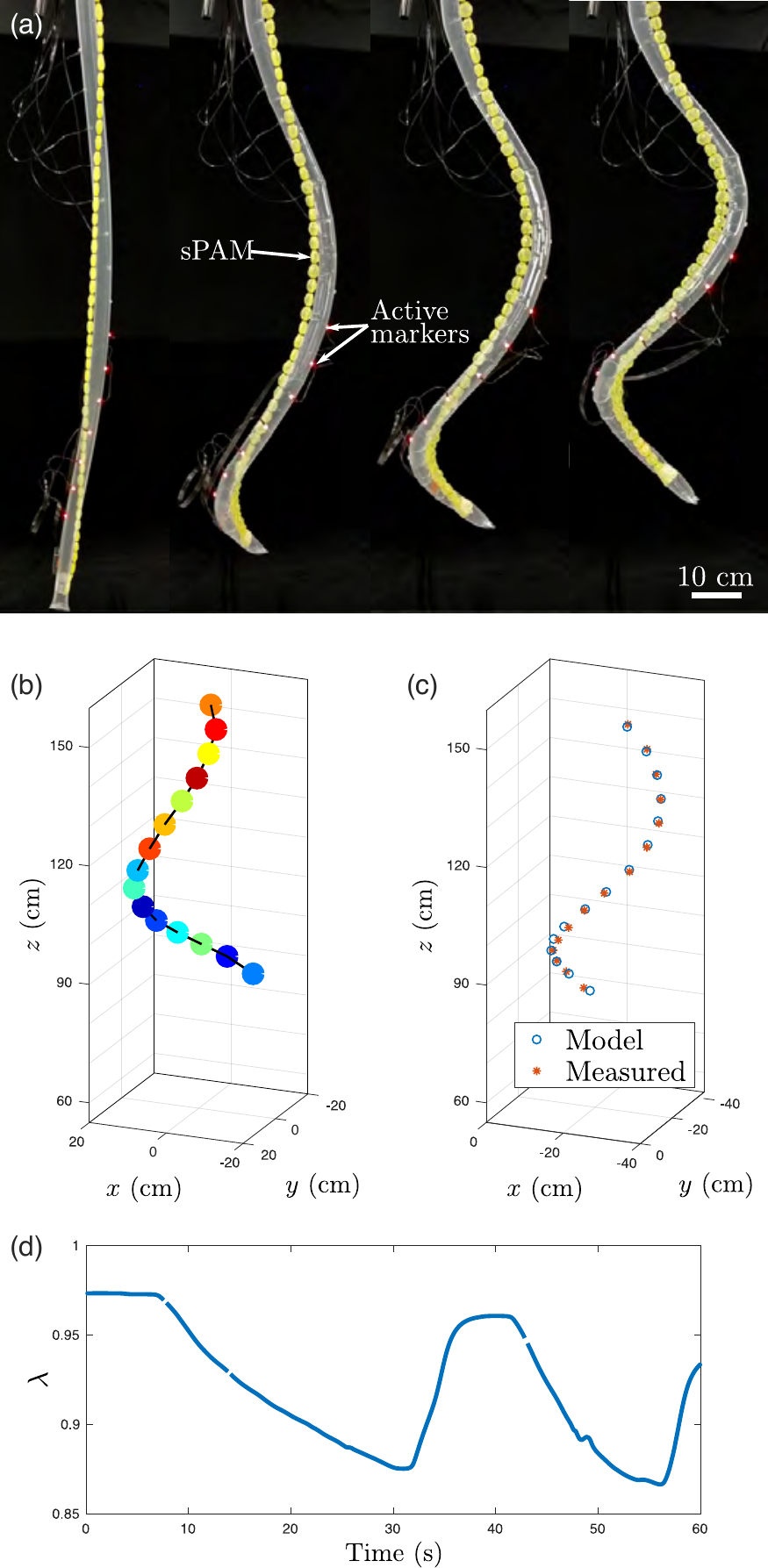}
	\caption{\small A series of actuation states for an active shape with a series pneumatic artificial muscle (sPAM) routed around a 3.23~cm diameter tube at an angle of $4^{\degree}$. The shape is measured by a set of active motion capture markers attached to the outer path of the helix. (a) The range of actuation demonstrated and the locations of the actuator and active markers. (b) A frame of the motion capture tracking showing the 15 tracked points. (c) Example fit between the model and the measured data. The optimization includes fitting $\lambda$ because it varies as the sPAM actuates. $R^2$ average is 0.96, with $\theta = 4^{\degree}$ and $D=3.23$~cm. (d) Variation of the fit $\lambda$ over two cycles of actuation. Gaps show where markers were obstructed or other errors caused the helix fitting to fail.}
	\label{fig:ch5_ActiveHelix}
\end{figure}

\begin{figure*}[t!]
\centering
	\includegraphics[width=1.9\columnwidth]{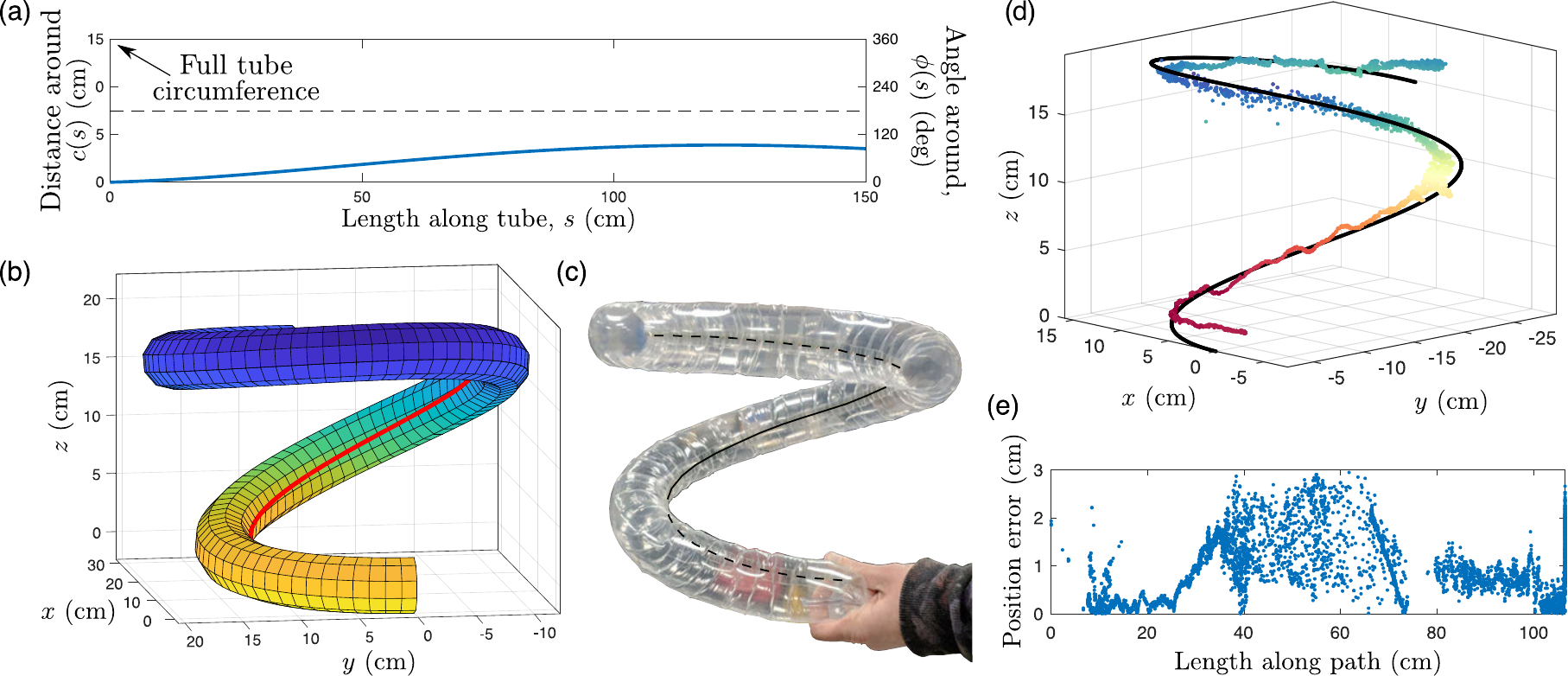}
	\caption[Actuator routing and the modeled and implemented shapes for a static general shape on a tube with diameter 4.77~cm.]{\small Actuator routing and the modeled and implemented shapes for a static general shape on a tube with diameter 4.77~cm. (a) Shape of the general actuator path drawn on the tube, generated from a polynomial function. The x-axis shows the length of tube and the y-axis shows the circumference of the tube. This allows represents the line of actuation drawn onto the tube when laid flat. (b) The predicted shape from the model given the actuator path in (a) and $\lambda$ equal to 0.7. The red line shows the actuator line on the deformed shape. (c) Resulting shape from implementing the actuator path in (a) on a physical static prototype. The black line shows the line of actuation. (d) Measured shape of the polynomial based actuator path aligned to the modeled shape using ICP. The measured points are colored based on the z-height. (e) Position error versus length along the path. The RMSE between the data set and model is 0.45~cm with a maximum error of 3.0~cm.}
	\label{fig:ch5_PolyyPathDesign}
\end{figure*}

\subsubsection{Actuated Helix}
We carried out a similar validation of the uniform helical actuator model using an active shape. Using an sPAM as the actuator (Section~\ref{subsec:PAMs}), we used a 3.23~cm diameter tube with a wall thickness of $68~\mu$m for the body and routed the sPAM at an angle of $4^{\degree}$. As the actuator shortens, the tube is shaped into a series of helices (Fig.~\ref{fig:ch5_ActiveHelix}(a)).

We measured the shape as the sPAM was actuated using a motion capture system with active LED markers (Impulse X2E Motion Capture, PhaseSpace). In total, 15 points along the outermost path of the helix were tracked (Fig.~\ref{fig:ch5_ActiveHelix}(b)). Since only two of the actuator parameters, $\theta$ and $D$, are static during actuation, we found the helix fit for the measured shape using the optimization above and adding $\lambda$ as a optimization variable (Fig.~\ref{fig:ch5_ActiveHelix}(c)). Due to the camera placement, partial or full marker occlusions occasional occurred during actuation. These frames were removed as they led to inaccurate shape matching (seen as small gaps in the plot in Fig.~\ref{fig:ch5_ActiveHelix}(d)). Overall, the fits had an average $R^2$ value of 0.96, on par with our fits for static shapes above. Fig.~\ref{fig:ch5_ActiveHelix}(d) shows the fit $\lambda$ values over time as the actuator is shortened and released for two cycles. The value of $\lambda$ ranges between 0.97 and 0.87, smaller than the range analyzed in the static helix experiments, but enough to create significant shape change as seen in Fig.~\ref{fig:ch5_ActiveHelix}(a). This range also matches well with the smallest achievable $\lambda$ of the sPAM measured with the actuator alone, $\lambda=0.73$, since we expect that the range of achievable $\lambda$ will be less when attached to a pressurized tube due to the stiffness of the pneumatic beam.

\subsection{General Actuation}
After validating the uniform helical actuation model from Section~\ref{subsec:ch5_UniformActModel}, we validate the extension of this model to continuously varying actuation (Section~\ref{subsec:ch5_GenActModel}). We again do this for both static and active actuation and extract the 3D path of the resulting shapes. For the general paths, we use the iterative closest point algorithm (ICP) to match the modeled and measured shapes \cite{besl1992method}. This works well for the general actuation since, unlike the uniform actuation, portions of the resulting shapes are not often self-similar to other portions of the shape, allowing the ICP algorithm to find a match more easily.

\subsubsection{Static General Path}

For the static path, we choose a routing based off a fourth order polynomial previously used in testing Cosserat-based models \cite{rucker2011}. The path routing is defined in terms of the angle of the actuator around the circumference, $\phi(s)$, parameterized in terms of the length along the tube, $s$:
\begin{equation}
\begin{multlined}
    \vspace{-0.3cm}
    \phi(s) = 5887\left(\frac{s}{1000}\right)^4-2847\left(\frac{s}{1000}\right)^3 \\\\
    +320\left(\frac{s}{1000}\right)^2+6\frac{s}{1000}
\end{multlined}
\end{equation}
where s is in centimeters, and $\phi(s)$ is in radians. The total length of the tube used, $\mathscr{L}$, is 150~cm. We plot this actuator path as it looks on the undeformed tube when flattened for construction purposes (Fig.~\ref{fig:ch5_PolyyPathDesign}(a)). This is done by calculating the distance of the path around the circumference, $c(s)=\frac{D}{2}\phi(s)$, and plotting it as a function of the length along the tube. The actuator parameter $\theta(s)$ can be calculated by rewriting Equation~(\ref{eqn:ch5_phi(s)}) as:
\begin{equation}
    \theta(s_i)=\arctan\left(\frac{D}{2}\frac{\phi(s_{i+1})-\phi(s_i)}{\Delta\ell(s_i)}\right).
\end{equation}
For the last actuator parameter, we choose $\lambda$ equal to 0.7. The resulting shape using the model in Table~\ref{table:solution} and the transformation matrices in Equations~(\ref{eq:ch5_transformationCC})-(\ref{eq:ch5_positionCC}) is shown in Fig.~\ref{fig:ch5_PolyyPathDesign}(b). Finally, the actuation is implemented in a physical prototype using a tube with diameter of 4.77~cm and wall thickness of $68~\mu$m, shown in Fig.~\ref{fig:ch5_PolyyPathDesign}(c). 
\begin{figure*}[tb!]
\centering
	\includegraphics[width=1.9\columnwidth]{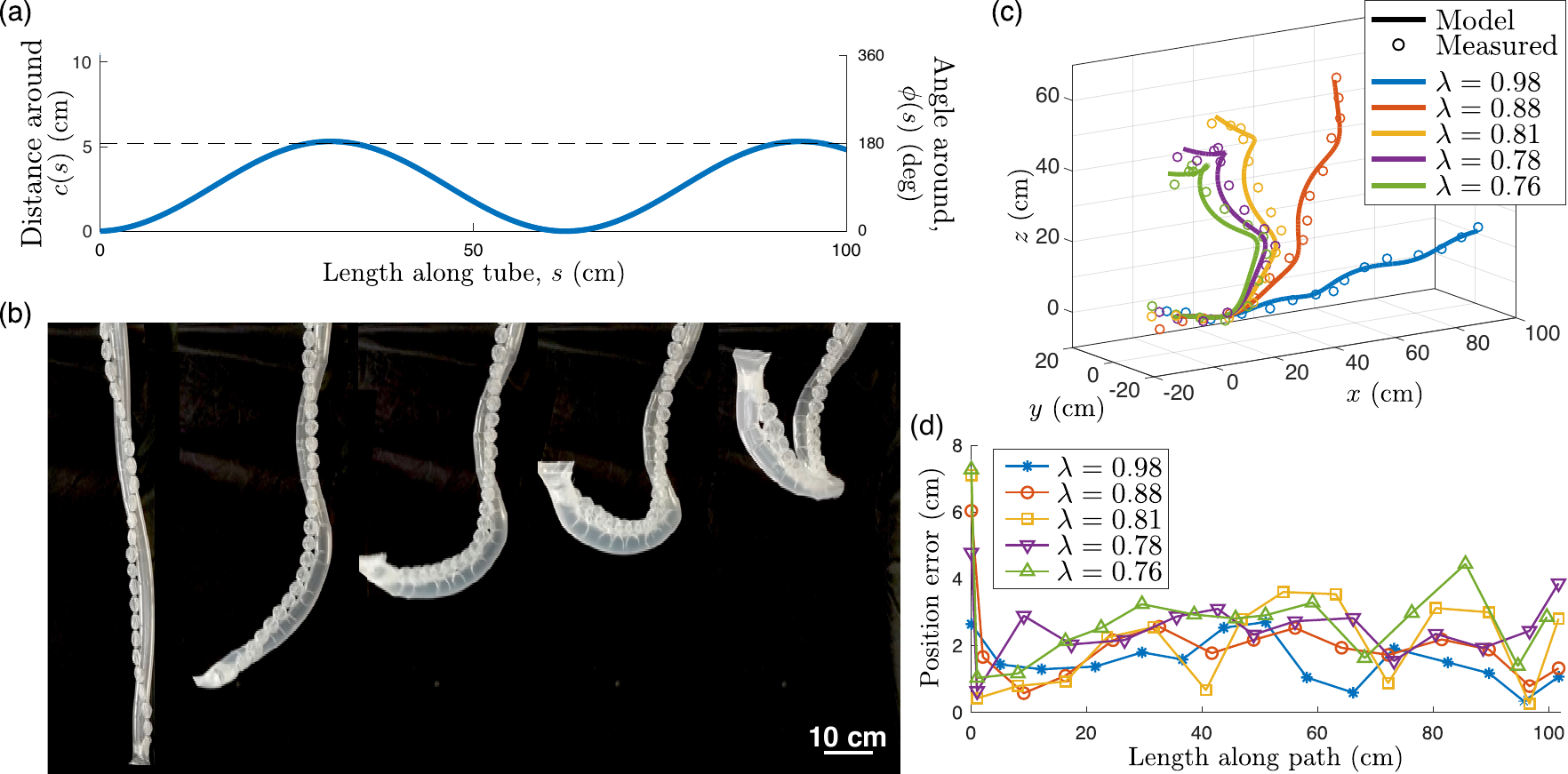}
	\caption{\small General actuation of a pneumatic backbone using an sPAM. The actuator routing is based on sinusoidally varying the value of $\theta$. (a) Shape of the general actuator with sinusoidal $\theta$ variation drawn on the tube as length along the tube versus the distance around the circumference. (b) Series of actuation states of the physical prototype with the actuator path from (a) as the sPAM is pressurized, changing $\lambda$. The sPAM had a minimum measured $\lambda$ value of 0.65 when actuated alone, and this range will be truncated by the stiffness of the soft growing robot body. (c) Measured and modeled shapes for the sPAM actuated generally routed path. The best fit $\lambda$ value for each measured shape, given the model, are used to generate the model comparison shape and are shown in the legend. (d) Position error versus length along the path for the sPAM actuated shapes. The measured shapes, with decreasing value of $\lambda$, have an RMSE equal to 1.67 cm, 2.36 cm, 2.88 cm, 2.72 cm, and 3.20 cm and maximum errors of 2.70~cm, 6.04~cm, 7.10~cm, 4.78~cm, and 7.27~cm.}
	\label{fig:ch5_TrigPathDesign}
\end{figure*}

Qualitatively, Fig.~\ref{fig:ch5_PolyyPathDesign} shows a good match between the modeled and implemented actuation. To quantify this, we measured the inner path of the implemented shape using the magnetic tracker. The magnetic tracker sensor was held up to the inner path and traced by hand. This path was aligned to the modeled path using the ICP algorithm. The results of the alignment are shown in Fig.~\ref{fig:ch5_PolyyPathDesign}(d) and the error over the length is shown in Fig.~\ref{fig:ch5_PolyyPathDesign}(e). The modeled shape and implemented shape match each other well, with an RMSE of 0.45~cm with a maximum error of 3.0~cm. 

\subsubsection{Actuated General Path}

We also implement an active generally routed actuator using sPAMs. Here we design $\theta(s)$ directly, choosing:
\begin{equation}
    \theta(s)=\frac{\pi}{12}\sin(0.1s)
\end{equation}
where $s\in[0,100]$~cm. The tube has a diameter of 3.31~cm with a wall thickness of $68~\mu$m and the length of each tube segment is $\Delta\ell(s)=1$~cm. The shape of the actuator around the circumference of the tube is calculated for implementation and shown in Fig.~\ref{fig:ch5_TrigPathDesign}(a). When the actuator is pressurized, the tube deforms into a series of non-uniform shapes with decreasing values of $\lambda$ (Fig.~\ref{fig:ch5_TrigPathDesign}(b)).

The shape is measured during actuation using the Impulse X2E Motion Capture system. Again, we track 15 points along the outer path of the shape. We then analyze a subset of the achieved shapes, shown in Fig.~\ref{fig:ch5_TrigPathDesign}(c), and attempt to minimize the RMSE from the ICP fit by varying the value of $\lambda$. Fig.~\ref{fig:ch5_TrigPathDesign}(c) shows the best fits and the values of $\lambda$ leading to those fits for five shapes along the full range of actuation and Fig.~\ref{fig:ch5_TrigPathDesign}(d) shows the position error as a function of length. The best fit $\lambda$ values range from 0.76 to 0.98. This $\lambda$ range is again confirmed by the measured maximum contraction of the actuator alone, $\lambda = 0.65$, taking into consideration the stiffness of the pneumatic backbone. The fits for the shapes, going from largest $\lambda$ to smallest, have RMSEs of 1.67~cm, 2.36~cm, 2.88~cm, 2.72~cm, 3.20~cm and maximum errors of 2.70~cm, 6.04~cm, 7.10~cm, 4.78~cm, and 7.27~cm. The error increases as the shape is actuated further, but these errors are still relatively small given the total length of the tube, 100~cm. This example also shows how even ``simple" tendon routings, like a sinusoid, can lead to rather complex final shapes.

\section{Inverse Design}
\label{sec:ch5_invDesign}
Section~\ref{sec:ch5_modelValidation} shows how the model can be used to accurately predict resulting shapes from a given actuator path, so we now demonstrate how we can use the model to design an actuator path that will actuate into a desired shape. While we have already shown the inverse model for helices in Table~\ref{table:solution}, the inverse solution is not trivial for general paths. In this section, we will discuss the actuator constraints and features we need to consider in the inverse design process, different methods for finding an actuator path to match a target shape, and demonstrations of the actuator design process.

\subsection{Design Constraints}
\label{subsec:ch5_constraints}
In switching focus from shape prediction to actuator design, the capabilities and limitations of each actuation method become important to consider. Some of these constraints apply to any style of actuation because they are more dependent on the magnitude and direction of contraction that is feasible to apply to the pneumatic backbone, while others are directly an impact of the type of actuator chosen. 

\begin{figure}[t!]
\centering
	\includegraphics[width=.8\columnwidth]{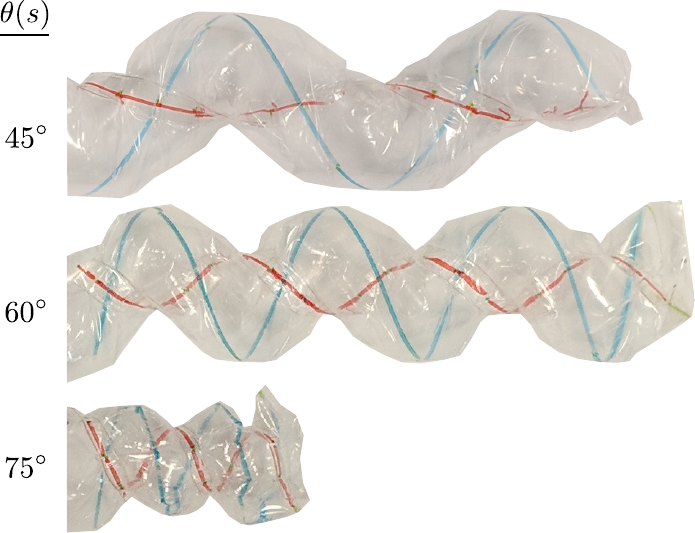}
	\caption{\small Helical shapes produced using high values of $\lambda$ to establish parameter limits. The actuator contraction ratio is kept at $\lambda = 0.5$. Angles over $60^{\degree}$ tend to produce poorly formed shapes without significant shape change that could be used to matching target shapes.}
	\label{fig:ch5_ThetaLimit}
\end{figure}

The first important constraint to consider is the bounds of the actuator parameters, $\theta$, $\lambda$, and $D$. While these are not absolute limits for the most part, they are useful practical limits. For the actuator angle, we look at the maximum angle that is feasible and/or useful for creating shapes. While we only tested up to $50^{\degree}$ for the uniform validation in Section~\ref{sec:ch5_modelValidation}, we tested larger values to find practical limits. The helices for $45^{\degree}$, $60^{\degree}$, and $75^{\degree}$ are shown in Fig.~\ref{fig:ch5_ThetaLimit}. While actuators with values of $\theta$ over $60^{\degree}$ can be physically constructed, we can see in the shape produced with $\theta=75^{\degree}$ that high values of $\theta$ lead to poorly formed helices and do not result in significant deviation from a straight tube, besides reducing the length and apparent diameter. While these effects might be useful in some cases, for the purposes of matching desired shapes we limit ourselves to $|\theta|\leq60^{\degree}$. For the limits of the contraction ratio, $\lambda$, we have one absolute limit: we can only allow actuator contraction, i.e.\ $\lambda\leq1$. This limit is based on the inextensibility of the pneumatic backbone materials. For the lower limit, we primarily consider the actuator capabilities to impose a parameter constraint. For example, depending on construction, the minimum $\lambda$ value for an unloaded sPAM is between 0.5 and 0.7 \cite{Greer2017}. Similarly, for mechanically programmed and tendon actuation, $\lambda$ values less than 0.4 were found to be difficult to produce consistently and impeded the growth function of the robot. These implementation constraints give practical lower limits for $\lambda$ depending on the implementation. Lastly, we consider the limits on diameter, $D$. While previous work has demonstrated soft growing robots over orders of magnitude in size \cite{hawkes2017}, consistent and precise construction of a tube of a desired diameter is not feasible. For this reason, unlike the other parameters where we limited the range, for the diameter we limit the choice to those available pre-manufactured off-the-shelf. This still provides us a wide range of choices, and does not limit the potential designs significantly, since diameter change only scales the produced shape \cite{blumenschein2018helical}.

The other important constraint to consider is the actuator implementation used, as well as any coupling between parameters due to that actuator implementation. As discussed in Section~\ref{sec:ch5_actImplementation}, the different implementations of tendon actuation lead to different trade-offs. Two of the implementations discussed, the mechanically programmed shapes and the tendon and stopper actuation, can only produce a single well-defined shape. However, we can easily implement paths where $\theta$ and $\lambda$ both vary with these actuator implementations. Pneumatic muscle actuation, as seen in Fig.~\ref{fig:ch5_ActiveHelix} and Fig.~\ref{fig:ch5_TrigPathDesign}, allows for a whole family of shapes to be actuated as the pressure changes, but it is difficult to vary $\lambda$ over the length of the actuator. For the pneumatic artificial muscles discussed, a single input pressure is used to control the actuator contraction, so any variations in $\lambda$ must be designed into the response of the pneumatic muscle to a given pressure. With an sPAM actuator, $\lambda$ can theoretically be varied by a small amount by changing the spacing of the o-rings \cite{Greer2017}. However it is not obvious if the ratio of $\lambda$ values will remain consistent as the shape is actuated. More modeling and testing would be needed to verify the feasibility of this strategy, so, for this design process, we limit ourselves to a single value of $\lambda$ over the actuator length when considering sPAM designs.

\subsection{Design Algorithm}

We developed an algorithm to iteratively design paths to match desired shapes. The algorithm is modified slightly depending on the actuator constraints described above, but each follows the same basic structure. We show the algorithm for the mechanically programmed implementation and tendon and stopper actuation in Algorithm~\ref{algorithm:design_stopper}. In the algorithm, we take the desired shape, $\vec{r}_{c,d}$, and find vectors of values for $\theta(s)$, $\lambda(s)$, and $\Delta\ell(s)$, as well as an initial transformation $\mathbf{T}^{0}_{w}$, that allow us to most closely match the shape. As discussed in Section~\ref{subsec:ch5_constraints}, we only have a few choices for the tube diameter, $D$, so we choose an appropriate value based on the size of the desired shape and give $D$ as an additional input to the algorithm. The actuator parameters are used to calculate the achieved path, $\vec{r}_c$, using Equation~(\ref{eqn:ch5_iterativeTransform}). We compare the desired and achieved shapes using the sum of the linear distances between each pair of points:
\begin{equation}
    e = \sum_{j} \|\vec{r}_{c}(j)-\vec{r}_{c,d}(j)\|
\end{equation}
where $e$ is the error in our shape matching. 

\begin{algorithm}[b!]
\caption{Tendon and Stopper Actuator Design Algorithm} 
\label{algorithm:design_stopper}
\begin{algorithmic}[1]
\footnotesize
\Function{Design}{$\vec{r}_{c,d},D,k,n$}
    \State ($\displaystyle \mathbf{T}^{0}_{w},\lambda_{1:k},\theta_{1:k},\ell_{1:k}) \gets \argminB_{\mathbf{T}^{0}_{w},\lambda,\theta,\ell}$ PathError($\vec{r}_{c,d},D,\mathbf{T}^{0}_{w},\lambda_{1:k},\theta_{1:k},\ell_{1:k}$)
    \State $N \gets $ length($\vec{r}_{c,d}$)
    \For{$j\gets 2$ to $N/n$}
        \State $m\gets (j+k-1)$
        \State ($\displaystyle \lambda_{j:m},\theta_{j:m},\ell_{j:m}) \gets \argminB_{\lambda,\theta,\ell}$ PathError($\vec{r}_{c,d},D,\mathbf{T}^{0}_{w},\lambda_{1:m},\theta_{1:m},\ell_{1:m}$) 
    \EndFor
\State \Return $\lambda,\theta,\ell$
\EndFunction
\\
\Function{PathError}{$\vec{r}_{c,d},D,\mathbf{T}^{0}_{w},\lambda,\theta,\ell$}
    \State $\vec{r}_c \gets$ actuatorToPath($D,\lambda,\theta,\ell$) 
    \State $\vec{r}_c \gets \mathbf{T}^{0}_{w}\vec{r}_c$ 
    \State $e \gets \sum_{i} \|\vec{r}_c(i)-\vec{r}_{c,d}(i)\|$ 
\State \Return $e$
\EndFunction

\end{algorithmic}
\end{algorithm}

\begin{figure*}[bt!]
\centering
	\includegraphics[width=1.9\columnwidth]{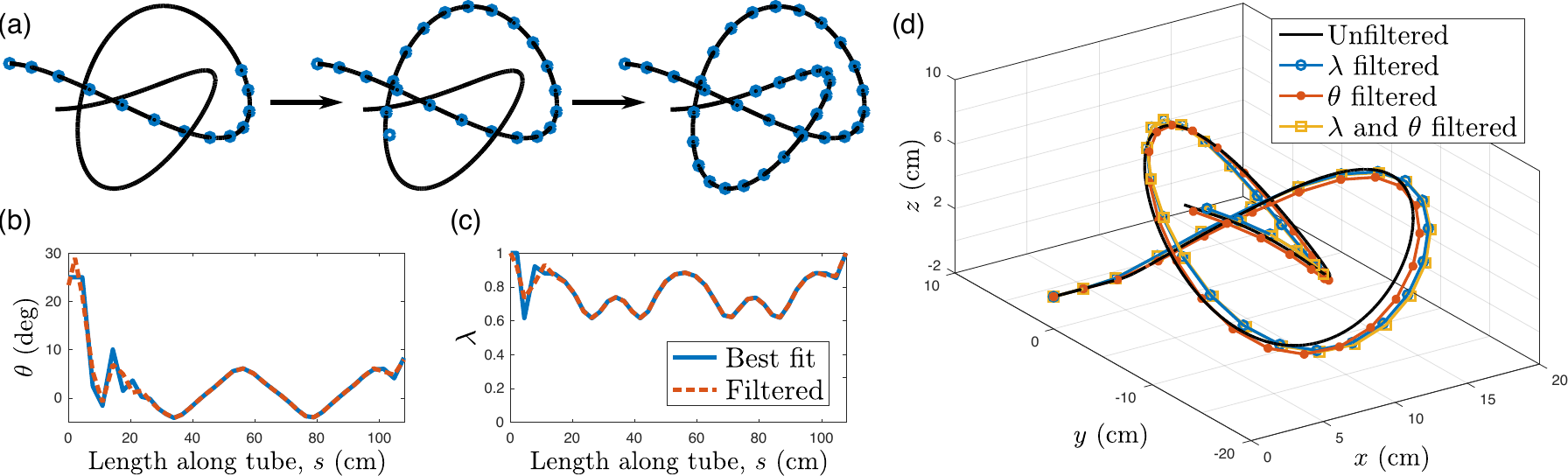}
	\caption[Example of the inverse actuator design process using tendon and stopper actuation to match a trefoil knot.]{\small Example of the inverse actuator design process using tendon and stopper actuation to match a trefoil knot. (a) Results of fitting to the trefoil knot using the design algorithm at three points in time. Each blue dot marks the end of an actuator segment covering $n=10$ points of the desired shape. The actuator segments are fit in order, with all previous segments held constant, and using a look-ahead fitting for the next $k=4$ segments. (b)-(c) The designed actuator parameters, $\lambda(s)$ and $\theta(s)$, along the length of the tube. These parameters can include high frequency noise, so they are low-pass filtered. (d) Plot showing the effects of filtering the actuator parameters.}
	\label{fig:ch5_TrefoilCurveFitting}
\end{figure*}

To find values for $\theta(s)$, $\lambda(s)$, $\ell(s)$, and $\mathbf{T}^{0}_{w}$ that produce a shape closely matching the target shape, we attempt to minimize the error, $e$. If we attempt to optimize over all the actuator parameters, with a set of parameters for each point along the path $\vec{r}_{c,d}$, the problem quickly becomes intractable as the number of variables to optimize grows. To lower the number of variables to optimize over, we took two approaches. First, we grouped sets of neighboring points into segments which will have the same actuator parameters, dividing the path in $\vec{r}_{c,d}$ into uniform helical segments with $n$ points each. Secondly, instead of optimizing all segments simultaneously, we treat the desired path, $\vec{r}_{c,d}$, like a trajectory. This means we optimize the actuator for each segment of the path given the fixed actuation state from the previous segments and considering the resulting error in the next $k$ segments. This look-ahead is important because the starting position and orientation of a segment is given by the previous actuation, so small errors can accumulate quickly if each segment is treated by itself. The results of this iterative process are described in Algorithm~\ref{algorithm:design_stopper} and can be seen in Fig.~\ref{fig:ch5_TrefoilCurveFitting}(a). Within Algorithm~\ref{algorithm:design_stopper}, line~2 fits the initial transformation and first $k$ helix parameters. The algorithm then iterates through the $N/n$ segments of the fit, with line~6 fitting $k$ helix parameters starting at the $j$th point. The function in line~12 is an implementation of the forward kinematics, Equation~(\ref{eqn:ch5_iterativeTransform}). In all cases, the optimization is performed in Matlab using a global search strategy with the \textit{fmincom} solver and \textit{sqp} algorithm. The global search is performed over $2000$ trial points for the initial segment and $300$ trial points for each additional segment. Initial values for the optimization variables for the first segment fit are set to $\lambda_{1:k,0}=0.75$, $\theta_{1:k,0}=0$, and $\ell_{j,0}=\|\vec{r}_{c,d}(s_{j+1})-\vec{r}_{c,d}(s_j)\|$. Subsequent segments use the best fits from the look ahead of previous segments to set the initial values.

For pneumatic artificial muscle implementation, we modified this algorithm slightly to account for the different constraint from the actuation: the constraint that we need a single $\lambda$ for the entire actuator. To incorporate this constraint, we no longer optimize the shape segments over $\lambda$. Instead, we give $\lambda$ as an additional input, like $D$, to the algorithm, and iterate through potential values of $\lambda$ external to the optimization. We save the actuator parameters and $\lambda$ value for the shape with the smallest error, $e$. Since this slows down our overall optimization, we speed up the search by stopping solutions that will clearly fail. For this criteria, we consider a design to have failed if the cost function reaches $nkD/2$, meaning on average each of the $nk$ points being analyzed at a step has an error equal to the tube radius.

\subsection{Demonstration}
\label{subsec:ch5_trefoilKnot}
\subsubsection{Actuator Design}
We demonstrate the inverse design solution against a series of target shapes. To start, we designed an actuator to allow a growing robot to tie a knot with its own body. We plan to implement this with tendon and stopper actuation, so we used the tendon actuation algorithm, Algorithm~\ref{algorithm:design_stopper}. The path chosen is a trefoil knot \cite{knotatlas} with the parameterization:

\begin{figure*}[bt!]
\centering
	\includegraphics[width=2\columnwidth]{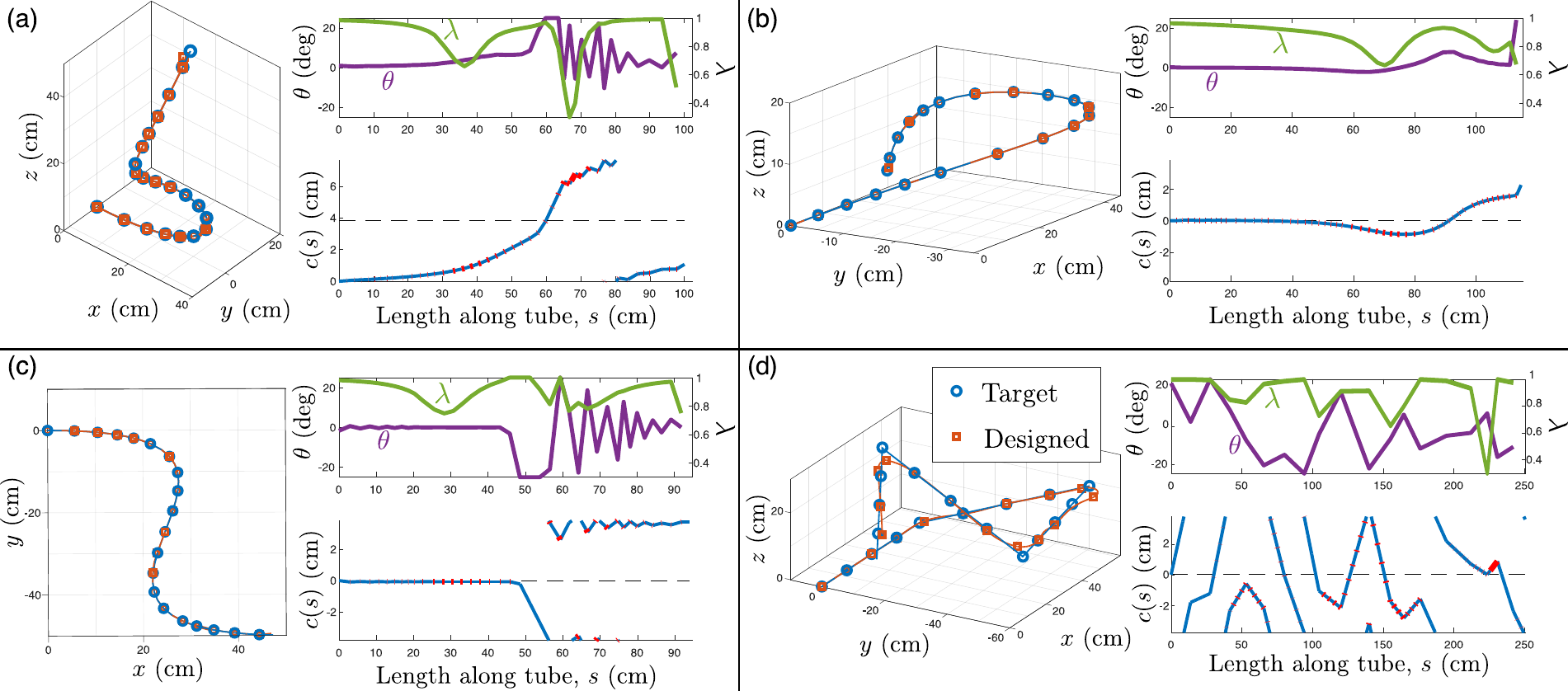}
	\caption{\small Examples of inverse actuator design process using tendon and stopper actuation showing 3D fit, actuator parameters over the length, and actuator shape. (a) Actuator fit for B\'ezier curve with control points  $[0,0,0]$, $[76,0,0]$, $[-4,19,42]$, $[34,-52,13]$, $[16,19,48]$, with RMSE 0.20~cm and maximum error 2.19~cm (b) Actuator fit for B\'ezier curve with control points  $[0,0,0]$, $[54,0,0]$, $[61,-50,18]$, $[8,-60,18]$, $[33,0,30]$, $[13,-10,6]$, with RMSE 0.13~cm and maximum error 1.22~cm. (c) Actuator fit for 2D B\'ezier curve with control points  $[0,0]$, $[50,0]$, $[24,-24]$, $[0,-50]$, $[50,-50]$, with RMSE 0.09~cm and maximum error 0.68~cm. (d) Actuator fit for a piecewise linear curve, interpolating between the control points of (b), with RMSE 2.23~cm and maximum error 7.20~cm.}
	\label{fig:beziercurves}
\end{figure*}

\begin{figure*}[bt!]
\centering
	\includegraphics[width=1.85\columnwidth]{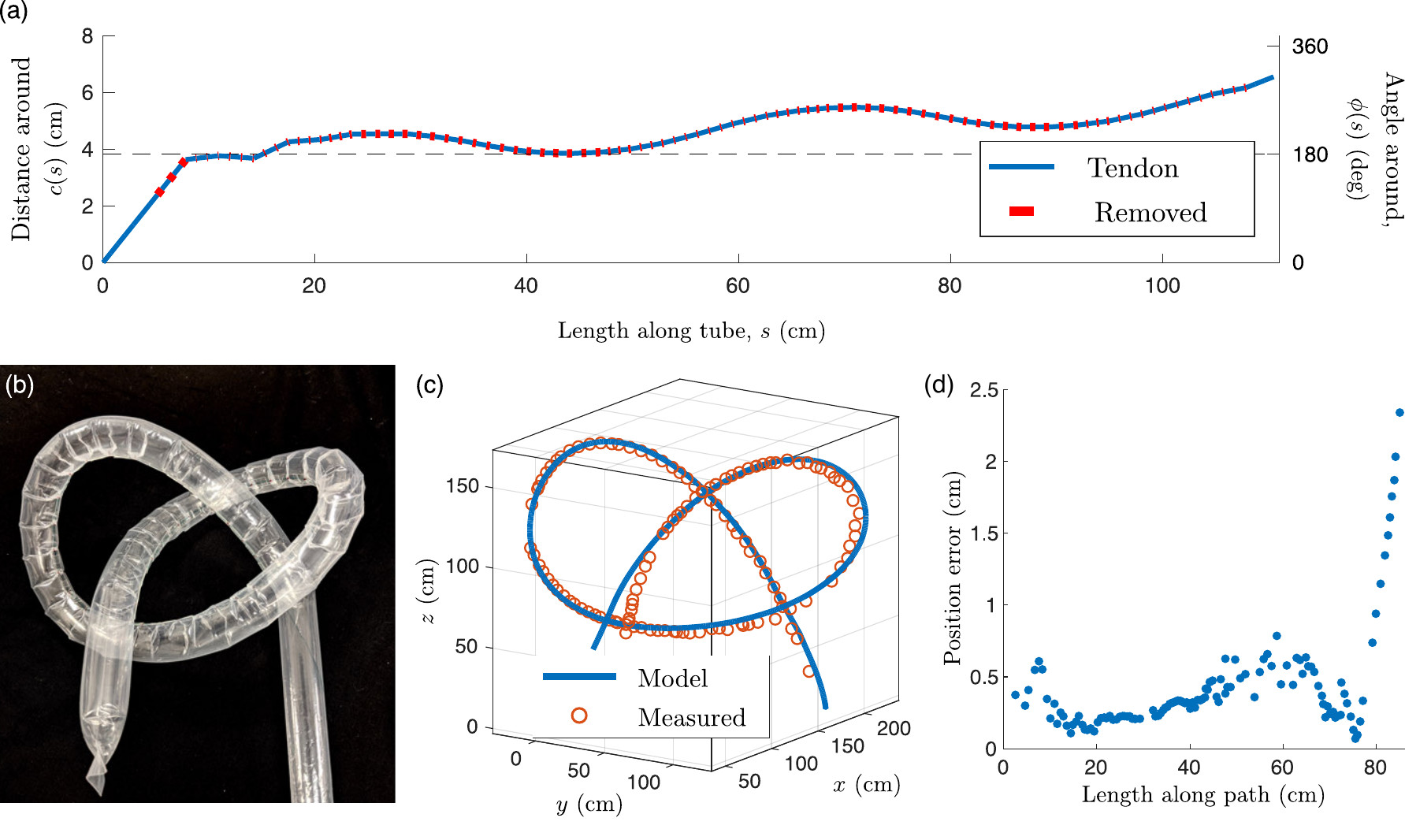}
	\caption[Design and implementation of the actuator path designed to achieve a trefoil knot.]{\small Implementation of the designed path to achieve a trefoil knot. (a) The shape of the actuator on a flat tube with diameter 2.43~cm. Blue represents the location of stoppers or unpinched areas and red shows where material is removed during actuation. (b) Physical implementation of the trefoil knot by mechanical programming. (c) Measurement of the trefoil knot in (b) using the magnetic tracker compared with the desired path. (d) Position error versus length along the path. The RMSE of the path is 6.88 mm and the maximum error is 25.3~mm.}
	\label{fig:ch5_TrefoilPathMeasure}
	\vspace{-0.5cm}
\end{figure*}

\begin{equation}
r(s)= \begin{bmatrix} m_x(\sin(t)+2\sin(2t))\\m_y(\cos(t)-2\cos(2t))\\ -m_z\sin(3t) \end{bmatrix} 
\label{eqn:ch5_knot}
\end{equation}
with $t\in[-0.2\pi,1.5\pi]$ and $m_x$, $m_y$, and $m_z$ scaling parameters for the shape. An example of this knot shape can be seen in Fig.~\ref{fig:ch5_TrefoilCurveFitting}. Using Algorithm~\ref{algorithm:design_stopper}, we designed actuators for a range of $125$ different trefoil knots for a tube of diameter 2.43~cm and a wall thickness of $68~\mu$m, varying the parameters $m_x$, $m_y$, and $m_z$ in Equation~(\ref{eqn:ch5_knot}) between 2 and 6. The paths were discretized into 400 points equally spaced in $t$. The algorithm was performed with $n=10$ and $k=4$, and it took an average time of $157$ seconds and a maximum time of $220$ seconds to complete on a laptop computer. The RMSE between the desired and modeled paths for the range of trefoil knots was $0.65$~mm on average and $1.57$~mm maximum over path lengths ranging between 620~mm and 1602~mm.  Fig.~\ref{fig:ch5_TrefoilCurveFitting}(b)-(c) shows the fit for the trefoil knot with $m_x=m_y=m_z=4$. For this scaling, the algorithm took $138$ seconds to complete and the RMSE between the desired and modeled paths was $0.22$~mm over a path length of 980~mm, with a maximum error of $0.48$~mm near the start. While for the most part the design algorithm produces actuator designs with smooth variation of the parameters, there are some higher frequency changes at the start and end of the path. This is likely due to those parts of the path having low curvature. For ease of construction, we decrease this parameter variation slightly by low-pass filtering the parameter curves. Since we target only the infrequent high frequency oscillations, this filtering does not have a major effect on the resulting shape, as shown in Fig.~\ref{fig:ch5_TrefoilCurveFitting}(d), and increases the RMSE relative to the desired path for the displayed trefoil knot to $\sim 0.76$~mm.

The algorithm performs similarly well on other smooth curves. Fig.~\ref{fig:beziercurves}(a)-(c) show the results of using Algorithm~\ref{algorithm:design_stopper} to fit tendon and stopper actuation paths for smooth B\'ezier curves, showing the fit $\lambda$ and $\theta$ values over the length as well as the resulting shape of the actuator on the tube. For each curve we used $n=10$, $k=4$, and 400 points along the length, resulting in an RMSE of 0.20~cm, 0.13~cm, and 0.09~cm relatively and taking an average of 199 seconds to calculate. While the algorithm performs well overall, the greedy nature of the algorithm leads to high oscillatory behavior in $\theta$ on straight sections in Fig.~\ref{fig:beziercurves}(a) and (c) due to overshooting the target shape. Interestingly, the algorithm is able to fit the in-plane s-curve seen in Fig.~\ref{fig:beziercurves}(c) very well. Lastly, we fit an actuator for a non-smooth curve with $n=10$, $k=4$, and 200 points along the length (Fig.~\ref{fig:beziercurves}(d)). The RMSE is higher, at 2.23~cm over the 250~cm length, but the shape manages to stay close to the target shape, with 1.64~cm of error at the end even after a maximum error of 7.20~cm is reached.

\subsubsection{Design Implementation}
We translate the actuator parameters from the trefoil knot in Fig.~\ref{fig:ch5_TrefoilCurveFitting} to an implementation plan using Equation~(\ref{eqn:ch5_phi(si)}) to calculate $\phi(s)$ for the actuator and using that to plot the distance around the circumference (Fig.~\ref{fig:ch5_TrefoilPathMeasure}(a)). In addition to the path of the actuator, we show the sections to be removed (in red). These sections are held pinched with tape in a mechanically programmed implementation or represent the segments without stoppers in the tendon and stopper implementation. The implementation using mechanical programming is shown in Fig.~\ref{fig:ch5_TrefoilPathMeasure}(b), and achieves the knot tying. We measure this shape with the magnetic tracker system and compare it to the desired path (Fig.~\ref{fig:ch5_TrefoilPathMeasure}(c)). The RMSE of the measured path relative to the designed path is 6.88~mm. The increase in error compared to the designed shape above is primarily due to manufacturing errors. Larger error between the measured and designed shapes develops at the end of the tube, with 25.3~mm maximum error at the tip. These points are at the end of the algorithm's fit sections and have relatively low actuation (as seen in Fig.~\ref{fig:ch5_TrefoilCurveFitting}(c)), so a small initial manufacturing error quickly grows without being corrected. This highlights the importance of the trajectory fitting method and suggests that future improvements to the design algorithm should consider the effects of manufacturing errors. We also implemented the actuation in Fig.~\ref{fig:ch5_TrefoilPathMeasure}(a) using a tendon and stopper actuator, which was then actuated by hand. The results of actuating the tendon pulled trefoil knot while growing can be seen in Fig.~\ref{fig:ch5_GrowVsNoGrow}(a)-(d).

\section{Discussion}
In the previous two sections, we have demonstrated using the geometric model with a piecewise helical formulation to both accurately predict an actuated shape given the shape of the actuator and to design actuators that can match desired shapes. Here we discuss conclusion drawn from the experimental results, the interaction of steering with growth, limitations within the produced shapes and model, and potential applications for shape matching.

\subsection{Forward and Inverse Model}
The static and active tests of the helical and general actuation show that the geometric model accurately predicts the resulting helical shape from the actuator parameters and that general paths can be approximated as a series of uniform helical paths with the correct alignment between segments. Areas where high amounts of error do exist are due to self-collision or occur near the ends of shapes. Best helical fits do not show any clear errors in the model shape or actuator parameters, additionally validating the model. Lastly, the active sPAM actuation further confirms the model by showing that continuously varying one of the actuator parameters, $\lambda$, over time leads to a family of shapes sharing the expected $\theta(s)$ and $D$ as predicted by the model.

The inverse design demonstrations show the ability to use this model as a design tool. Smooth and non-smooth 3D curves can be matched and challenging 2D s-curves can be achieved with single actuators. Improvements in the optimization algorithm could lead to quicker optimization of designs with finer discretization. The examples do show areas for improvement, including the tendency to create high oscillations in parameters, which will be addressed in future~work.

\begin{figure*}[tb!]
\centering
	\includegraphics[width=1.80\columnwidth]{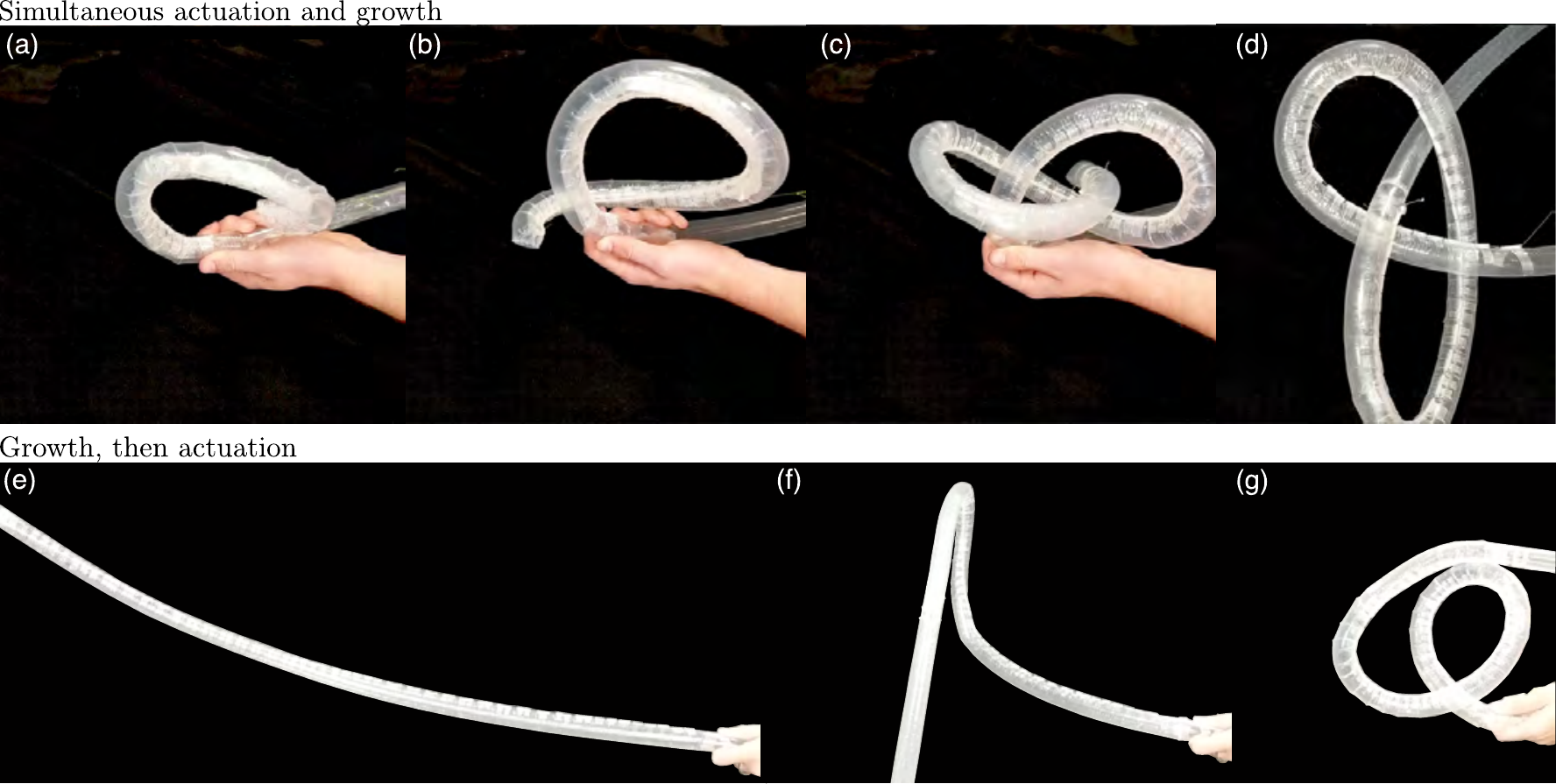}
	\caption[Implementation of the trefoil knot path using a tendon and stopper actuator. Top row shows successful knot tying with growth and actuation together, bottom row shows failed knot tying by using actuation after growth.]{\small Implementation of the trefoil knot path using a tendon and stopper actuator pulled by hand. The top row shows successful knot tying with growth and actuation together. The bottom row shows failed knot tying by using actuation after growth. (a)-(d) The robot is simultaneously actuated and grown. The knot requires two self crossings to form; (b) shows the first self crossing, and (c) shows the second when the knot is tied. (d) Tension in the tendon is released and the robot remains knotted. (e)-(g) The robot is actuated after being grown out. (f) The tube sweeps through a large area as the actuator is shortened. (g) The robot fails to knot when fully actuated due to self-collision.}
	\label{fig:ch5_GrowVsNoGrow}
\end{figure*}

\subsection{Actuation and Growth}
While the previous experiments and demonstrations focus on the shape change, the thin walled body of the inflated-beam robot is specifically chosen to allow for ``growth" from the tip through material eversion \cite{hawkes2017}, and this ability to grow adds interesting and useful features when it comes to general shape actuation. These features can primarily be broken down into two ideas: expanding the types of shapes that can be made with a single actuator, and expanding the environments where these shapes can be actuated and used. 

Both of these features can be seen in Fig.~\ref{fig:ch5_GrowVsNoGrow}, where we compare actuating a tendon driven trefoil knot while it is growing versus actuating after the tube is fully extended. First, growth allows us to actuate into a knot with a single actuator. We can see two points in the growing sequence, shown in Fig.~\ref{fig:ch5_GrowVsNoGrow}(b)-(c), when the path crosses itself, allowing the knot to form and to stay tied in Fig.~\ref{fig:ch5_GrowVsNoGrow}(d). Comparing this to the actuation without growth, Fig.~\ref{fig:ch5_GrowVsNoGrow}(g) shows the fully actuated tube in self-collision at the two locations, which are the two crossing points of the knot, so the knot is unable to form. This demonstrates that growing while actuating allows the system to create features and geometries that cannot be achieved normally with a single actuator alone by taking advantage of growing to move around potential self-collisions. In addition, we can imagine exploiting self-collision in the function of the actuation, like in Fig.~\ref{fig:ch5_GrowVsNoGrow}(d), where the self-collisions that were avoided using growth are now active as the tendon relaxes, allowing the robot to stay tied. 

Using growth during the actuation also reduces the amount of free space needed to produce the shape, allowing us to avoid unnecessary collision with the environment. The grown path remains contained to the perimeter of the knot until we are done actuating in Fig.~\ref{fig:ch5_GrowVsNoGrow}(a)-(d), while the directly actuated path in Fig.~\ref{fig:ch5_GrowVsNoGrow}(e)-(f) swings through a wide area around the tube. When actuating in a constrained space, it may be difficult or impossible to fully actuate a non-growing robot, meaning growing will be the only way to fully deploy the goal shape.

Growing while actuating does have downsides as well. As the total angle swept out by a path increases, the force to actuate the tendon and the pressure needed to grow both increase as an exponential function of that angle \cite{blumenschein2017modeling}. Thus, growing into a path will become more difficult the further along the path the robot grows, and may actually not be possible depending on the pressure available. We found periodically relaxing and re-tensioning the tendon was found to encourage growth to continue in Fig.~\ref{fig:ch5_GrowVsNoGrow}(a)-(d). It is possible that by releasing the tendon and lowering the curvature slightly, we could avoid stopping the growth or restart it when it did stop. This friction is the result of internal material contacting the inflated tube, which previous work has mitigated by carrying the material at the tip \cite{haggerty2019characterizing} or by switching to a material with lower self-friction, like some coated fabrics~\cite{NaclerioRAL2020}.

\subsection{Actuator and Shape Limitations}

The infeasibility of large $\theta$ values places some limits on the types of curves that can be achieved. Creating a planar curve that instantaneously switches between positive and negative curvatures, i.e.\ an s-curve, requires the actuator to route at a high angle, ideally a $90^{\degree}$ angle, for a short section, as seen in the design in Fig.~\ref{fig:beziercurves}(c). This is not possible with the pneumatic artificial muscle actuators, and it is also potentially difficult to manufacture and actuate high angles for any implementation, as shown in Fig.~\ref{fig:ch5_ThetaLimit}. Put another way, the $\theta$ limit and the requirement that the actuator shape be continuous along the length place a hard limit on the instantaneous torsion values that can develop between segments of the actuator, though we will investigate relaxing these restrictions in the future for mechanically programmed implementations.

\subsection{External Forces and Buckling}
In this model, we ignore forces and instead focus on the geometry of the problem to generate the kinematics of the tendon actuation. While we showed that this method allowed us to generate accurate predictions of shape, it also ignores external forces, which may cause anywhere from small errors in the shape to large localized errors when local buckling of the membrane causes large movement of the structure. While large deformation under gravity or other forces was not observed in the tests preformed, previous work has shown that individual pre-programmed turns cause locations where the robot will preferentially bend when growing into obstacles \cite{greer2019robust}, so it is possible that local wrinkling along the actuator length also leads to lower resistance to deformation. More investigation is needed on how shape actuation affects the stiffness of the pneumatic robot backbone and how external forces affect the shape, in order to predict and correct for these errors.

\subsection{Applications}

The general shape actuation of a soft growing robot described here can be used to create desired deployable shapes. This model allows us to make a wide range of shapes that can be deployed from a small initial size, and that can be deployed where space restrictions exist. Deployable structures could be used for support or to exert forces on the environment, acting as a structural beam or a pneumatic jack \cite{hawkes2017}. Shape control also allows for navigation through delicate environments without exerting forces on the environment that may be undesirable. If the precise shape of a path in space is known, we can design an actuator or mechanically program the robot to grow into the path, a technique previously proposed for creating growing catheters \cite{li2021vine,berthet2021mammobot}. Lastly, shape control could be used to create complex movements at the tip of the soft growing robot to grab objects or interact with the environment in a specific way. We can design actuation to wrap the robot around an object to be manipulated, or even use the self-tying knot actuation to give the robot support in its environment as it is moving, and the strength of the robot body material in tension would allow for large pulling forces. Future work will look at the allowable errors in shape for different applications to better judge the results of the model and actuator design method.

\section{Conclusion}

In this work, we showed how a single actuator placed on a soft pneumatic continuum robot can create a large range of 3D shapes. We introduced a model for general actuator routings that built off our previous model for uniform helical routings. By validating this model, we showed that general actuations can be treated as a series of uniform actuations linked together. In addition to predicting the actuated shapes of the robot, we showed that this model can be used to design general actuators to create desired shapes, including tying the robot into a knot.

In the future, we want to design faster and more accurate algorithms for finding the optimal actuator routing for a desired path. The algorithms described in this work were unlikely to find the optimal actuator overall, since fitting segments sequentially results in a locally greedy strategy. Future algorithms may be able to find more optimal actuators by refitting earlier segments occasionally. We are also interested in creating design algorithms that target different parts of the resulting actuation, i.e.\ matching a desired movement of the tip of the robot, or matching multiple target shapes with a single actuator. In addition, we want to expand the model to cover more actuation situations, including understanding the interaction of multiple actuators and adding external loading effects to the model. While we have shown the wide range of shapes a single actuator can achieve, multiple actuators will allow for an even larger design space and, by accounting for external forces, the model can help design actuation strategies for carrying a payload or interacting with obstacles.

\appendices




\ifCLASSOPTIONcaptionsoff
  \newpage
\fi



%
\bibliographystyle{IEEEtran}
\bibliography{IEEEabrv,bare_jrnl}

%

\begin{IEEEbiography}[{\includegraphics[width=1in,height=1.25in,clip,keepaspectratio]{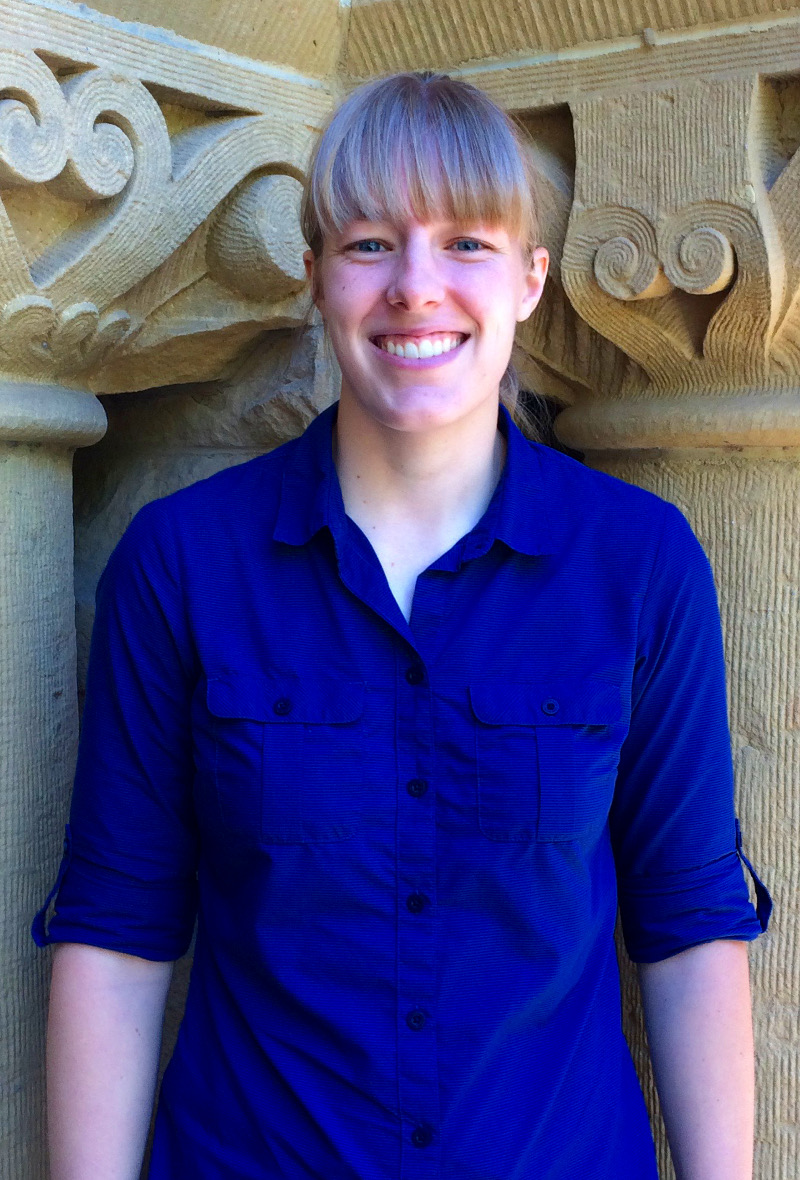}}]{Laura H. Blumenschein}
(Member, IEEE) received the B.S. and M.S. degrees from Rice University, Houston, TX, USA in 2015 and 2016, respectively, and the Ph.D. degree from Stanford University, Stanford, CA, USA in 2019 under the supervision of Prof. A. Okamura, all in mechanical engineering.

She is an Assistant Professor of Mechanical Engineering at Purdue University. Her research interests include soft robotics, actuator design, modeling, haptics, and growing robots.
\end{IEEEbiography}
\vspace{-1cm}
\begin{IEEEbiography}[{\includegraphics[width=1in,height=1.25in,clip,keepaspectratio]{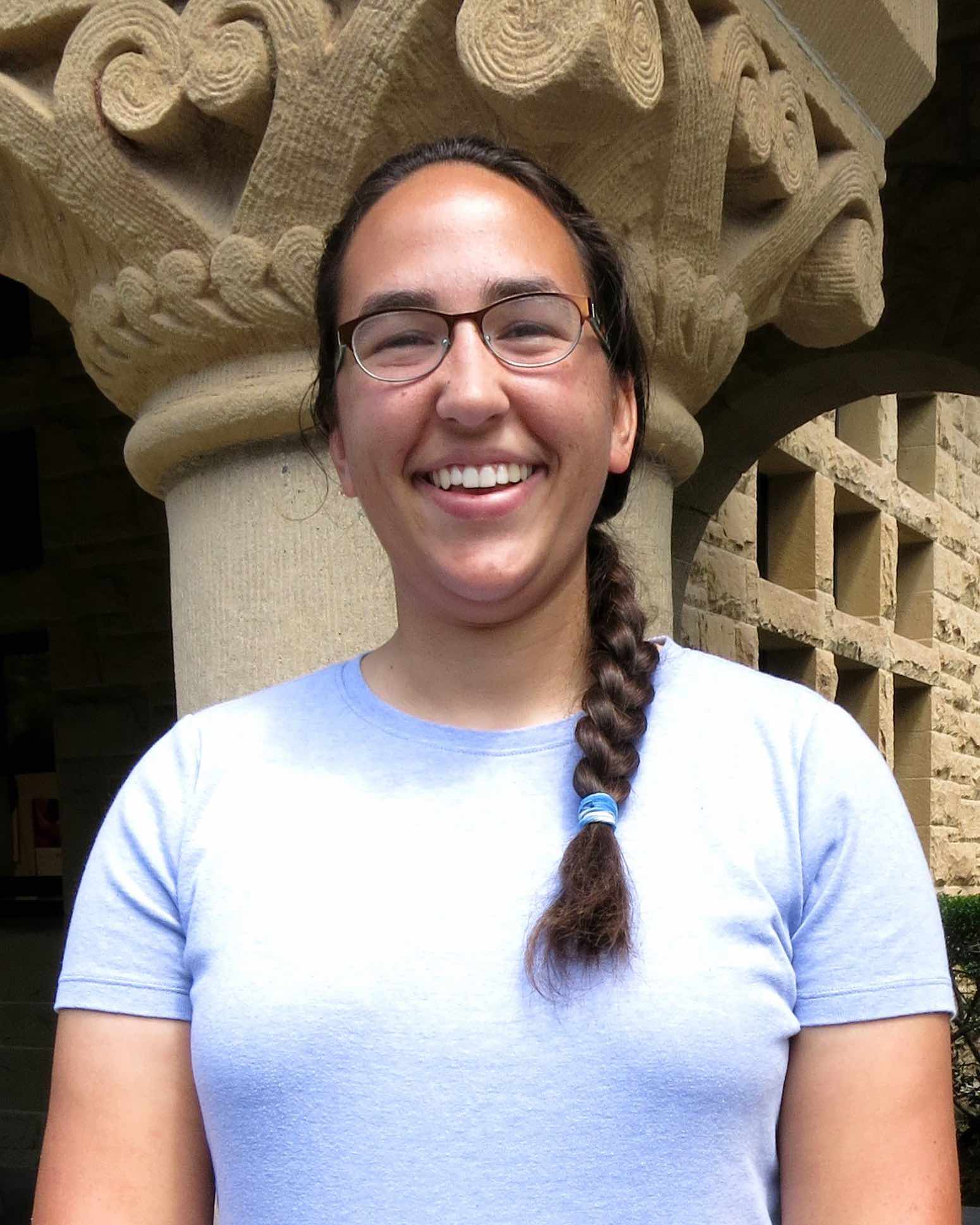}}]{Margaret Koehler}
(Student Member, IEEE) received the B.S., M.S., and Ph.D.\ degrees from Stanford University, Stanford, CA, USA, in 2014, 2016, and 2020, respectively, all in mechanical engineering. 

 She is now at Intuitive Surgical, Inc. Her research interests at Stanford University included robotics, haptics, automated design, and computational modeling.

\end{IEEEbiography}
\vspace{-1cm}
\begin{IEEEbiography}[{\includegraphics[width=1in,height=1.25in,clip,keepaspectratio]{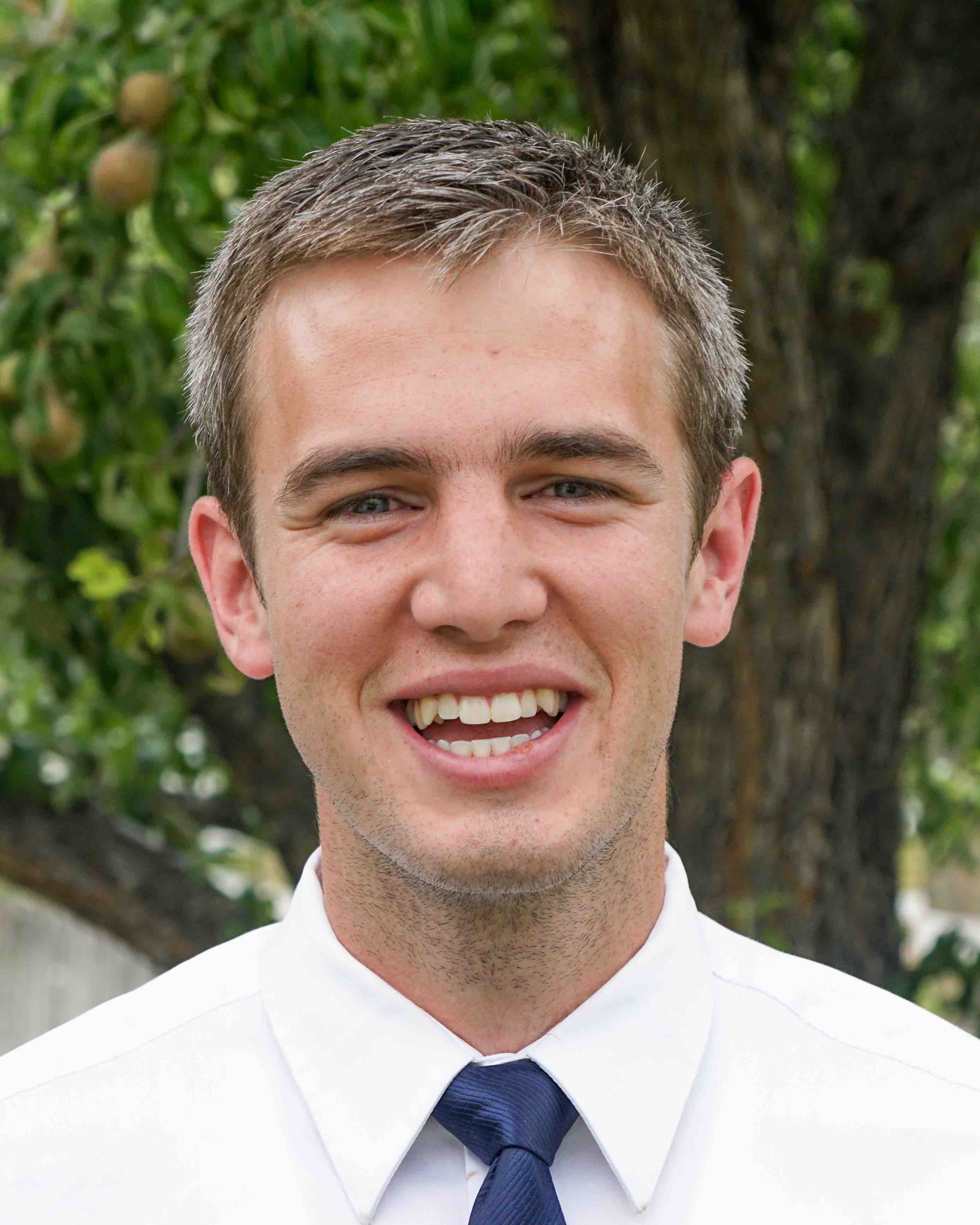}}]{Nathan S. Usevitch}
(Student Member, IEEE) received the B.S. degree from Brigham Young University, Provo, UT, USA in 2015 and the M.S.\ and Ph.D.\ degrees from Stanford University, Stanford, CA, USA, in 2017 and 2020, respectively, all in mechanical engineering.

He is now at Facebook Reality Labs. His research interests at Stanford University included actuator design, soft haptic devices, and soft robotics.
\end{IEEEbiography}
\vspace{-1cm}

\begin{IEEEbiography}[{\includegraphics[width=1in,height=1.25in,clip,keepaspectratio]{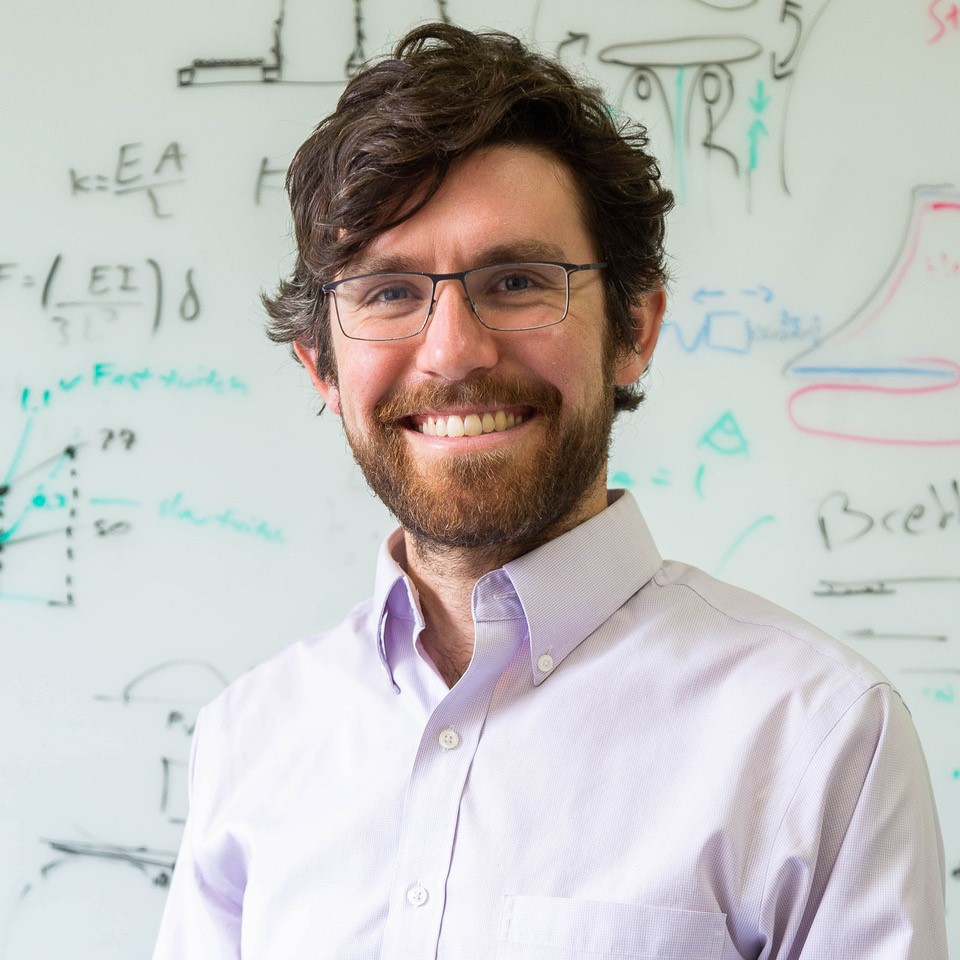}}]{Elliot W. Hawkes}
(Member, IEEE) received the A.B. degree with highest honors in mechanical engineering from Harvard University, Cambridge, MA, USA; the M.S. degree in mechanical engineering from Stanford University, Stanford, CA, USA; and the Ph.D. degree in mechanical engineering from Stanford University under the supervision of Prof. M. Cutkosky, in 2009, 2012, and 2015, respectively.

He is an Assistant Professor with the Department of Mechanical Engineering, University of California, Santa Barbara, Santa Barbara, CA, USA. His research interests include compliant robot body design, mechanism design, nontraditional materials, artificial muscles, directional adhesion, and growing robots. Dr. Hawkes received the NSF CAREER award in 2020.
\end{IEEEbiography}
\vspace{-1cm}
\begin{IEEEbiography}[{\includegraphics[width=1in,height=1.25in,clip,keepaspectratio]{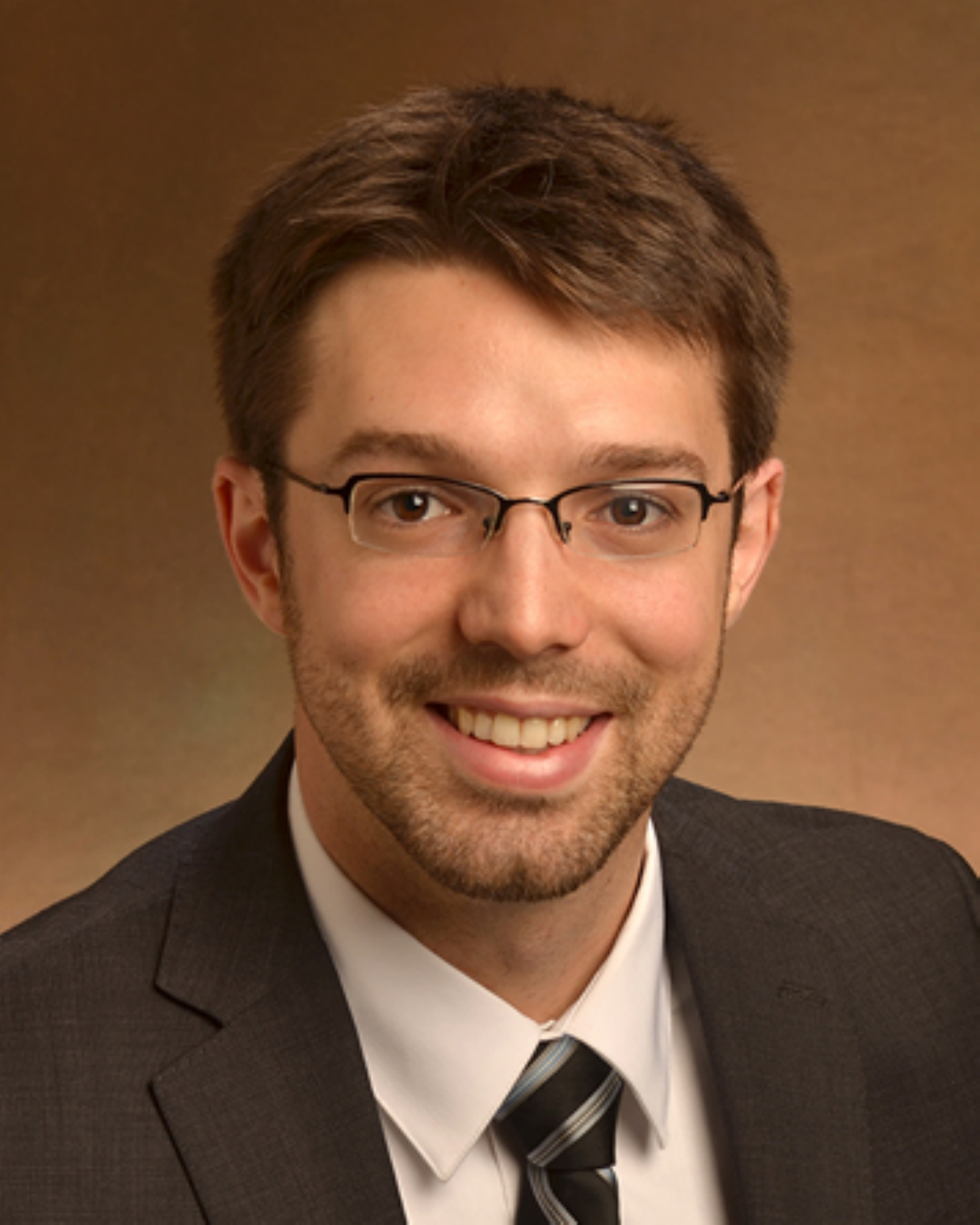}}]{D. Caleb Rucker} (Member, IEEE)
received the B.S. degree in engineering mechanics and mathematics from Lipscomb University, Nashville, TN, USA, in 2006, and the Ph.D. degree in mechanical engineering from Vanderbilt University, Nashville, in 2011.
He is currently an Associate Professor of Mechanical Engineering at The University of Tennessee, Knoxville, TN, USA, where he directs the Robotics, Engineering, and Continuum Mechanics in Healthcare Laboratory (REACH Lab).
Dr. Rucker was a recipient of the NSF CAREER Award in 2017.
\end{IEEEbiography}
\vspace{-1cm}
\begin{IEEEbiography}[{\includegraphics[width=1in,height=1.25in,clip,keepaspectratio]{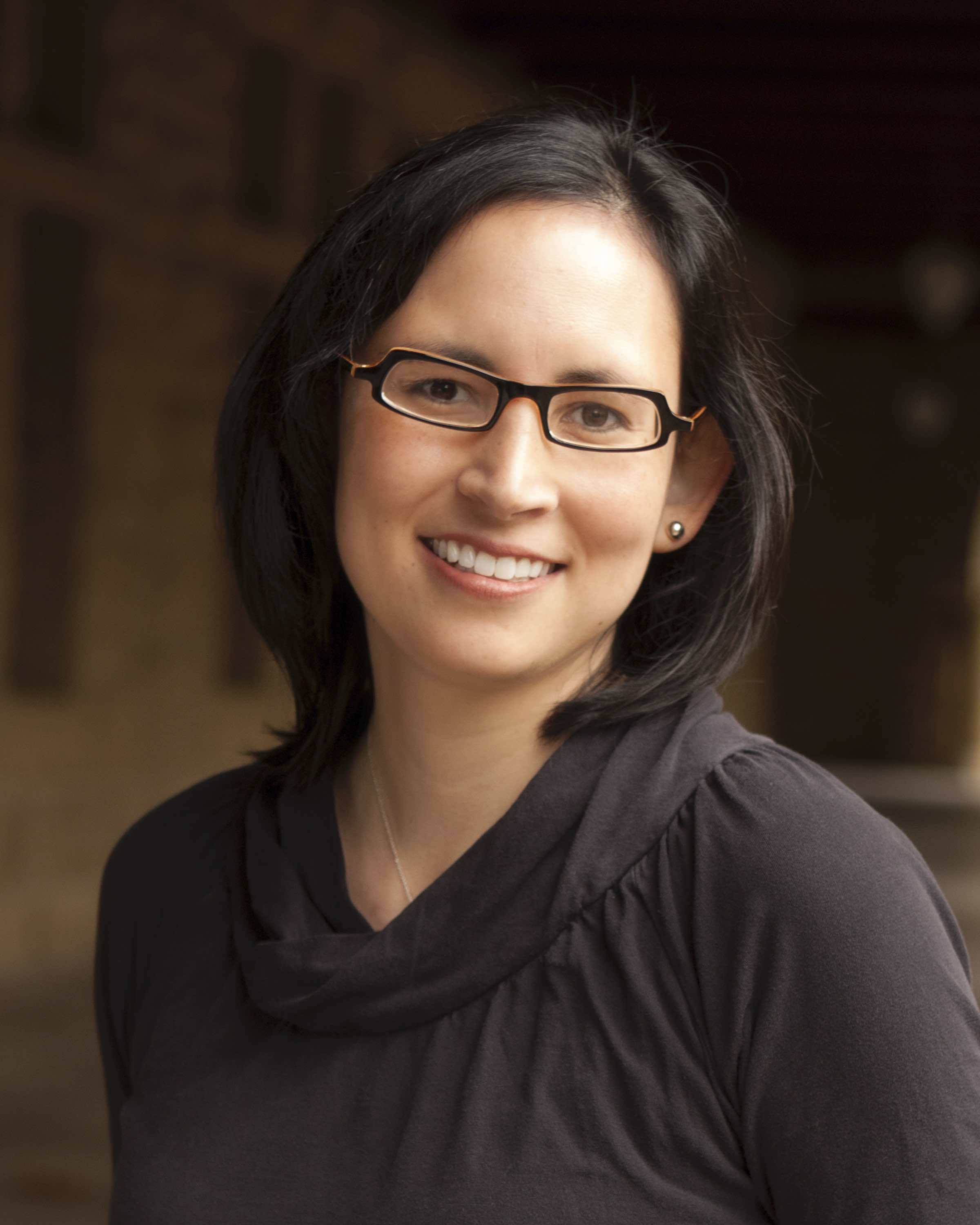}}]{Allison M. Okamura}
(Fellow, IEEE) received the B.S. degree from the University of California, Berkeley, Berkeley, CA, USA, in 1994, and the M.S. and Ph.D. degrees from Stanford University, Stanford, CA, USA, in 1996 and 2000, respectively, all in mechanical engineering.\\
She is currently a Professor of Mechanical Engineering with Stanford University. Her research interests include haptics, teleoperation, medical robotics, virtual environments and simulation, neuromechanics and rehabilitation, prosthetics, and engineering education. Prof. Okamura was the Editor-in-Chief of the IEEE ROBOTICS AND AUTOMATION LETTERS from 2018-2021.
\end{IEEEbiography}


\vfill


\end{document}